\documentclass{bmvc2k}
\usepackage{booktabs}
\usepackage{makecell}
\usepackage{graphicx}
\usepackage{float}
\usepackage{graphicx}
\usepackage{xcolor}
\usepackage{tikz}
\usepackage{amsmath,amssymb}
\usepackage{algorithm}
\usepackage{algpseudocode}
\usetikzlibrary{positioning,fit,backgrounds,calc}
\usepackage{pifont}
\usepackage{changepage}

\newcommand{\cmark}{\ding{51}}
\newcommand{\xmark}{\ding{55}}

\title{RefDiffNet: Learning to Expose Subtle PCB Defects Before Detection}

\addauthor{Vinay Edula}{vinaykumarreddyedula@gmail.com}{1}
\addauthor{Nilesh Badwe}{nbadwe@iitk.ac.in}{2}
\addauthor{Priyanka Bagade}{pbagade@iitk.ac.in}{1}

\addinstitution{
 Department of Computer Science and Engineering\\
 Indian Institute of Technology Kanpur\\
 Kanpur, India
}
\addinstitution{
 Department of Materials Science and Engineering\\
 Indian Institute of Technology Kanpur\\
 Kanpur, India
}

\runninghead{RefDiffNet}{Learning to Expose Subtle PCB Defects Before Detection}

\begin{document}
\maketitle

\begin{abstract}
Printed circuit board (PCB) defect detection is challenging because many defects are small and difficult to distinguish from complex background patterns. Most deep learning-based PCB inspection methods rely only on the inspected PCB image for defect detection, ignoring the defect-free reference image that encodes the expected layout of traces, pads, and other PCB structures. In this work, we propose \textbf{RefDiffNet}, a lightweight plug-and-play input enhancement block placed before the detector backbone to enhance the image before defect detection. RefDiffNet brings one proven idea from classical inspection into the deep learning era, using a defect-free reference image to reveal defects. RefDiffNet compares the defective image with the aligned reference, captures structural changes relative to the reference, and uses a lightweight encoder to output the original image with defective regions highlighted, thereby making the downstream detector's task easier. Results on HRIPCB and DeepPCB show that RefDiffNet consistently improves performance across detector families, including one-stage detectors from YOLOv8 to YOLOv26, the transformer-based RT-DETR, and the two-stage Faster R-CNN. It achieves up to $18\%$ relative mAP$_{50:95}$ gain with negligible overhead, introducing only $0.004$--$0.005$M additional parameters and $0.7$--$0.8$ GFLOPs, amounting to at most $0.25\%$ of the parameter count of any evaluated detector. Results establish RefDiffNet as a lightweight, plug-and-play, detector-agnostic input enhancement module that substantially improves PCB defect detection with minimal computational cost.

\end{abstract}


\begin{adjustwidth}{0.15in}{0.15in}

\section{Introduction}
\label{sec:intro_related}
Printed circuit boards (PCBs) form the fundamental backbone of modern electronic systems, providing mechanical stability and electrical connectivity for various components. The rapid advancement of AI and high-performance computing is driving the electronics industry toward extreme hardware miniaturization. This shift demands multilayer high-density interconnect (HDI) PCBs with microscopic features, including trace widths of $\leq 100 \mu m$ and microvias with diameters of $< 150 \mu m$. Fabricating these intricate structures requires sophisticated techniques with rigorous control over materials, processing conditions, and tooling. Even a single micro-defect introduced   

\end{adjustwidth}

during bare board manufacturing can catastrophically compromise the board's electrical integrity, rendering it non-functional long before any components are mounted. Because these sub-millimeter anomalies are imperceptible to the human eye, manual inspection is fundamentally impractical in high-volume manufacturing environments. Consequently, automated PCB defect detection has become an indispensable quality-control mandate in modern electronics fabrication~\cite{Ling2023Survey,chensurvey}. 


Early PCB inspection methods commonly relied on template image-based comparison, where an inspected board image was matched against a defect-free reference image to locate differences. A tested PCB image was aligned with a defect-free reference image, and defects were detected from the difference between the two images~\cite{Ling2023Survey,He2025Review}. Classical methods used image subtraction, thresholding, logical operations such as XOR, connected-component analysis, and morphology-based post-processing to isolate defective regions~\cite{Chauhan2011Subtraction,Putera2010Morphology}. Other methods compared wavelet responses, especially Haar sub-bands, to capture structural changes in traces and pads more robustly than raw pixel differencing~\cite{Ibrahim2005Wavelet}. More recent reference-based approaches such as ChangeChip also compare inspected and golden PCB images by constructing local difference descriptors and clustering mismatch patterns~\cite{changechip}. These methods are simple and efficient, and they directly exploit the reference image. However, they are sensitive to illumination variation, noise, small misalignment, which limit their robustness across different boards and imaging conditions~\cite{Ling2023Survey,He2025Review}.

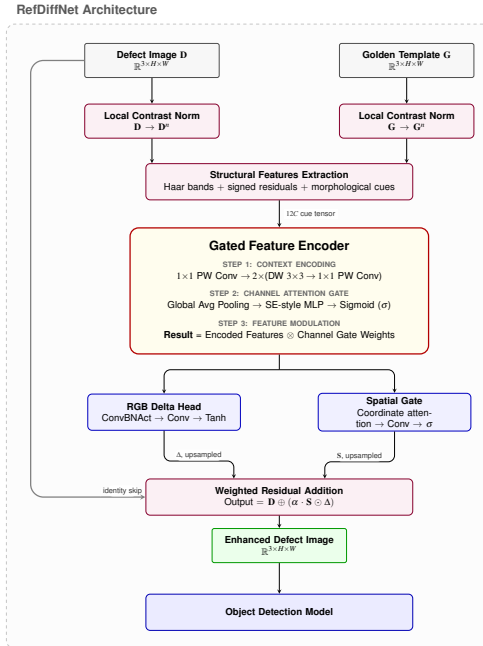
\begin{figure}[H]
    \centering
    \resizebox{0.50\columnwidth}{!}{%
    \begin{tikzpicture}[
        >=stealth,
        thick,
        font=\sffamily,
        tensor/.style={
            rectangle, draw=black!70, fill=gray!6, rounded corners=2pt,
            minimum height=3.2em, text width=4.2cm, align=center, font=\sffamily\small
        },
        op/.style={
            rectangle, draw=purple!70!black, fill=purple!6, rounded corners=4pt,
            minimum height=3.2em, text width=4.2cm, align=center, font=\sffamily\small
        },
        mergeop/.style={
            rectangle, draw=purple!70!black, fill=purple!6, rounded corners=4pt,
            minimum height=3.4em, minimum width=8.8cm, text width=8.4cm, align=center, font=\sffamily\small
        },
        encoder/.style={
            rectangle, draw=red!70!black, fill=yellow!8, rounded corners=8pt,
            line width=1.2pt, inner sep=12pt, text width=9.0cm, align=center, font=\sffamily\small
        },
        head/.style={
            rectangle, draw=blue!75!black, fill=blue!6, rounded corners=4pt,
            minimum height=3.4em, text width=4.8cm, align=center, font=\sffamily\small
        },
        math/.style={
            circle, draw=black!70, fill=white, inner sep=0pt,
            minimum size=0.75cm, font=\bfseries\large, align=center
        },
        label/.style={
            font=\sffamily\scriptsize, text=black!80, fill=white, inner sep=2pt
        },
        imgNode/.style={
            inner sep=0pt, draw=black!50, line width=1pt, rounded corners=2pt, overflow=hidden
        }
    ]


    \node[tensor] (defect) at (-4.2, 0)
    {\textbf{Defect Image} $\mathbf{D}$\\$\mathbb{R}^{3\times H\times W}$};
    
    \node[tensor] (golden) at (4.2, 0)
    {\textbf{Golden Template} $\mathbf{G}$\\$\mathbb{R}^{3\times H\times W}$};

    \node[op] (lcn_d) at (-4.2, -2.0)
    {\textbf{Local Contrast Norm}\\$\mathbf{D}\rightarrow \mathbf{D}^{n}$};
    
    \node[op] (lcn_g) at (4.2, -2.0)
    {\textbf{Local Contrast Norm}\\$\mathbf{G}\rightarrow \mathbf{G}^{n}$};

    \node[mergeop] (cues) at (0, -4.0)
    {\textbf{Structural Features Extraction}\\
    Haar bands $+$ signed residuals $+$ morphological cues};

    \node[encoder] (enc_gate) at (0, -7.6) {
        {\large \textbf{Gated Feature Encoder}}\\[1.5ex]
        \begin{tabular}{c}
            \textcolor{black!60}{\scriptsize \textbf{STEP 1: CONTEXT ENCODING}} \\
            $1{\times}1$ PW Conv $\rightarrow 2{\times}$(DW $3{\times}3 \rightarrow 1{\times}1$ PW Conv) \\[1.2ex]
            \textcolor{black!60}{\scriptsize \textbf{STEP 2: CHANNEL ATTENTION GATE}} \\
            Global Avg Pooling $\rightarrow$ SE-style MLP $\rightarrow$ Sigmoid ($\sigma$) \\[1.2ex]
            \textcolor{black!60}{\scriptsize \textbf{STEP 3: FEATURE MODULATION}} \\
            \textbf{Result} = Encoded Features $\otimes$ Channel Gate Weights
        \end{tabular}
    };

    \node[head] (rgb) at (-3.8, -11.6)
    {\textbf{RGB Delta Head}\\
    ConvBNAct $\rightarrow$ Conv $\rightarrow$ Tanh};
    
    \node[head] (spatial) at (3.8, -11.6)
    {\textbf{Spatial Gate}\\
    Coordinate attention $\rightarrow$ Conv $\rightarrow \sigma$};

    \node[mergeop, text width=6.8cm] (combiner) at (0, -14.4)
    {\textbf{Weighted Residual Addition}\\
    Output $= \mathbf{D} \oplus (\alpha \cdot \mathbf{S} \odot \Delta)$};

    \node[tensor, fill=green!8, draw=green!60!black] (output) at (0, -16.0)
    {\textbf{Enhanced Defect Image}\\$\mathbb{R}^{3\times H\times W}$};

    \node[mergeop, fill=blue!8, draw=blue!70!black, text width=6cm] (obj_det) at (0, -18.2)
    {\textbf{Object Detection Model}};

    
    \draw[->] (defect.south) -- (lcn_d.north);
    \draw[->] (golden.south) -- (lcn_g.north);

    \draw[->] (lcn_d.south) -- (lcn_d.south |- cues.north);
    \draw[->] (lcn_g.south) -- (lcn_g.south |- cues.north);

    \draw[->] (cues.south) -- node[right=4pt, label] {$12C$ cue tensor} (enc_gate.north);

    \coordinate (split_feat) at (0, -10.2);
    \draw[-] (enc_gate.south) -- (split_feat);
    \draw[->, rounded corners=6pt] (split_feat) -| (rgb.north);
    \draw[->, rounded corners=6pt] (split_feat) -| (spatial.north);

    \draw[->, rounded corners=6pt] 
        (rgb.south) -- ++(0,-1.0) -| 
        node[pos=0.25, label, above] {$\Delta$, upsampled} 
        ([xshift=-1.5cm]combiner.north);
        
    \draw[->, rounded corners=6pt] 
        (spatial.south) -- ++(0,-1.0) -| 
        node[pos=0.25, label, above] {$\mathbf{S}$, upsampled} 
        ([xshift=1.5cm]combiner.north);

    \draw[->] (combiner.south) -- (output.north);

    \draw[->] (output.south) -- (obj_det.north);

    \draw[->, rounded corners=12pt, color=black!50, line width=1.2pt]
        (defect.west) -- 
        (-8.2, 0) -- 
        (-8.2, -14.4) --
        node[pos=0.8, label, above] {identity skip}
        (combiner.west);

    \begin{scope}[on background layer]
        \coordinate (top_left) at (-9.0, 1.2);
        \coordinate (bottom_right) at (7.0, -19.4);

        \node[
            rectangle, draw=gray!45, dashed, fill=gray!2, rounded corners=8pt,
            fit=(top_left) (bottom_right), inner sep=0pt
        ] (box) {};

        \node[anchor=south west, font=\sffamily\bfseries\large, text=black!65, yshift=6pt, xshift=6pt] 
        at (box.north west) {RefDiffNet Architecture};
    \end{scope}

    \end{tikzpicture}
    }
    \caption{RefDiffNet}
    \label{fig:refdiffnet_gated_unified}
\end{figure}

Deep learning has significantly improved PCB defect detection by learning data-driven features instead of relying on hand-crafted rules. General object detectors such as Faster R-CNN~\cite{fasterrcnn} and modern YOLO families~\cite{yolo8,yolo9,yolo10,yolo11,yolo12,yolo26_ultralytics} have been widely adapted for PCB defect detection. Recent works improve backbone, neck, attention modules, feature fusion, or bounding-box loss to better handle tiny defects and complex circuit backgrounds. Notable methods include deep context learning with enhanced feature pyramids and refined CIoU loss~\cite{Lim2023DeepContext}, DDTR with CNN and Swin Transformer feature fusion~\cite{Feng2023DDTR}, CDI-YOLO with coordinate attention and depthwise convolution~\cite{cdiyolo}, YOLOv8-DEE with efficient multi-scale attention and EIoU loss~\cite{Yi2024YOLOv8DEE}, YOLO-BFRV with BiFPN-based feature fusion~\cite{yolobfrv}, EffNet-PCB for lightweight YOLOv8-based detection~\cite{Hou2025EffNetPCB}, improved YOLOv12 variants~\cite{improvedyolo12}, Lite-DETR~\cite{Luo2025LiteDETR}, and MSA-DETR for multi-scale alignment and spatial calibration~\cite{Zhang2025MSADETR}. Although these methods are effective, they usually depend only on the defective image and leave the reference image unused, even though it provides rich information for identifying deviations from the expected PCB layout.

Reference-guided deep learning methods address this limitation by using both the inspected image and the corresponding reference image. Tang \emph{et al.} introduced the DeepPCB dataset and a paired-image detection framework, where the template and tested images are jointly processed to predict defect categories and locations~\cite{deeppcb}. The HRIPCB dataset further supports evaluation under high-resolution PCB inspection settings where small and subtle defects are common~\cite{hripcb}. Other reference-guided methods mainly use the reference image to verify whether an inspected region is defective, rather than directly detecting defect locations. Auto-VRS first compares the inspected PCB region with the corresponding defect-free reference region after the initial automated optical inspection (AOI) stage, and then uses a neural network to distinguish real defects from false alarms~\cite{Deng2018AutoVRS}. Siamese-network-based methods compare an inspected patch with its corresponding defect-free reference patch to detect mismatches without assuming a fixed set of defect categories~\cite{Ding2020}, while CSS-Net uses cost-sensitive Siamese learning for true-defect and pseudo-defect classification after AOI~\cite{Miao2021CSSNet}. Reference image-guided methods also extend to PCB assembly inspection, where the reference board is used to check whether components are missing, misplaced, or incorrect~\cite{DVQI2023} and Siamese semantic segmentation for welding defect detection~\cite{DeepSiameseWelding2022}. These works show the value of reference information, but they usually use the paired images to make the final prediction directly, estimate similarity between image patches, or verify AOI alarms.

In this paper, we propose \textbf{RefDiffNet}, a lightweight reference-guided enhancement block that acts as a plug-and-play pre-backbone module before an off-the-shelf object detector. Given a defective PCB image and its aligned reference non-defective image, RefDiffNet first applies local contrast normalization to reduce photometric variation, then constructs Haar-band structural maps, signed defective-reference difference maps, and morphology-expanded mismatch maps. A lightweight gated encoder combines these cues and predicts an RGB correction map, which is added back to the defective image to make defect regions more visible before detection. In this way, RefDiffNet brings together the strengths of classical template-based inspection and the learning capability of modern deep detectors, while avoiding hand-crafted thresholds and without replacing or modifying the downstream detector architecture.

The main contributions of this work are:
\begin{itemize}
    \item We propose \textbf{RefDiffNet}, a lightweight plug-and-play pre-backbone enhancement module that uses an aligned reference image to highlight PCB defect regions before detection, without modifying the downstream detector architecture.
    
    \item We design complementary defective-reference cues, including Haar-band structural responses, signed residual maps, and morphology-expanded residual cues, and fuse them through a lightweight gated encoder to generate a spatially selective RGB enhancement.

\item We show that RefDiffNet generalizes across detector families on HRIPCB and DeepPCB~\cite{hripcb,deeppcb}, improving one-stage YOLOv8--YOLOv26 detectors, transformer-based RT-DETR, and two-stage Faster R-CNN with negligible overhead of only $0.004$--$0.005$M parameters and $0.7$--$0.8$ GFLOPs, corresponding to less than $0.25\%$ of the parameter count of even the smallest evaluated detector.

    \item The ablation studies show the contribution of each cue type and the gated enhancement design, confirming that reference-guided input enhancement is an effective strategy for PCB defect detection.
\end{itemize}

\section{Methodology}
\label{sec:methodology}

\subsection{Overview}

The proposed RefDiffNet is a lightweight plug-and-play pre-backbone enhancement module for PCB defect detection as illustrated in Figure~\ref{fig:refdiffnet_gated_unified}. Given a defective image $I_d \in \mathbb{R}^{C \times H \times W}$ and its aligned golden reference image $I_g \in \mathbb{R}^{C \times H \times W}$, RefDiffNet extracts reference image-guided structural and residual cues, predicts an RGB correction map, and applies it through a spatial gate before passing the enhanced image to the detector. In production environments where the inspection and golden reference images may not be perfectly aligned, a standard image registration step such as Enhanced Correlation Coefficient (ECC) maximization~\cite{Evangelidis2008ECC} can be applied before RefDiffNet to align the reference image with the defective image.

The enhanced image is computed as
\begin{equation}
    I_{\mathrm{out}} = I_d + \alpha \, S \odot \Delta,
    \label{eq:refdiffnet_output}
\end{equation}
where $\Delta \in \mathbb{R}^{C \times H \times W}$ is the learned RGB correction with respect to $I_g$, $S \in \mathbb{R}^{1 \times H \times W}$ is the learned spatial gate, $\odot$ denotes element-wise multiplication with broadcasting across channels, and $\alpha$ is a learnable scalar initialized to a small value. Thus, $\Delta$ determines what correction to add, while $S$ determines where the correction should be applied.

\subsection{Reference Image-Guided Cue Construction}

RefDiffNet first constructs a compact cue tensor from the defective-reference pair ($I_d, I_g$). These cues include local structural responses from Haar bands, signed residual differences, and morphology-expanded residual maps.

\begin{figure}[H]
    \centering
    \includegraphics[width=\textwidth]{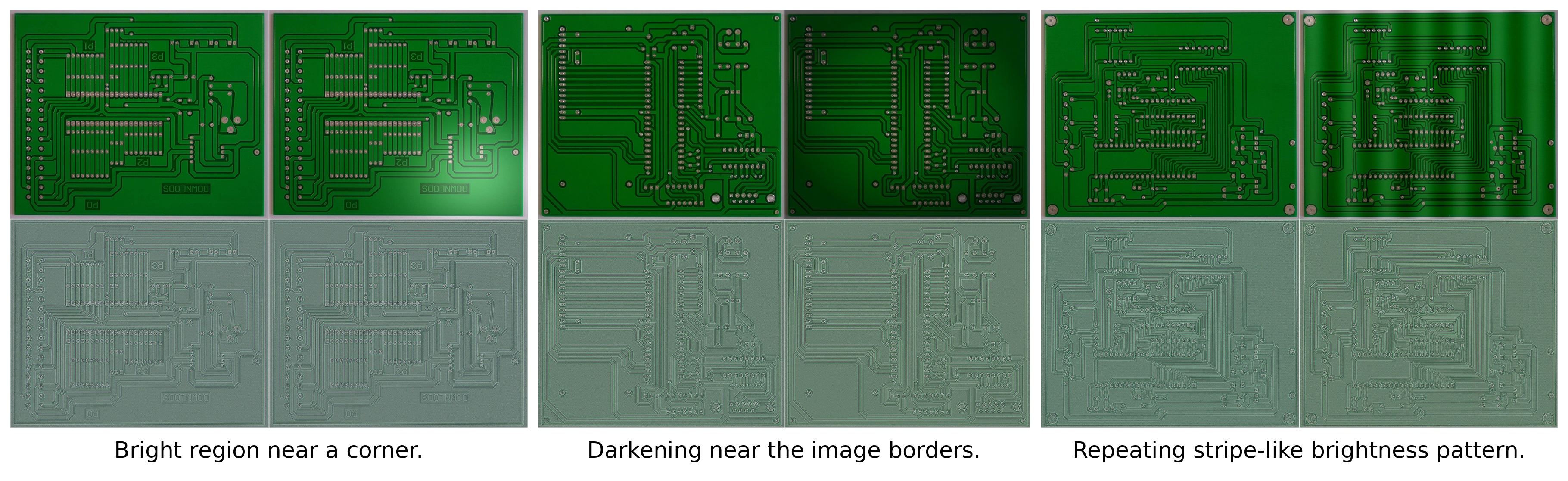}
    \caption{Effect of local contrast normalization under lighting changes. Each $2 \times 2$ grid compares the original and lighting-modified images with their LCN-normalized versions.
}
    \label{fig:lcn_qualitative_examples}
\end{figure}

\paragraph{Local contrast normalization.}
To reduce illumination and local contrast mismatch between the defective ($I_d$) and reference ($I_g$)images, both images are normalized before comparison:
\begin{equation}
    \hat{I}_d = \mathrm{LCN}(I_d), \qquad
    \hat{I}_g = \mathrm{LCN}(I_g).
\end{equation}
For each channel, local contrast normalization (LCN)~\cite{lcn} is applied using a $7 \times 7$ neighborhood $\Omega_{u,v}$:
\begin{equation}
    \mathrm{LCN}(I)(u,v) =
    \frac{I(u,v)-\mu_{u,v}}{\sqrt{\sigma_{u,v}^{2}+\epsilon}},
    \label{eq:lcn}
\end{equation}
where
\begin{equation}
    \mu_{u,v} =
    \frac{1}{|\Omega_{u,v}|}
    \sum_{(i,j)\in\Omega_{u,v}} I(i,j),
    \qquad
    \sigma_{u,v}^{2} =
    \frac{1}{|\Omega_{u,v}|}
    \sum_{(i,j)\in\Omega_{u,v}}
    \left(I(i,j)-\mu_{u,v}\right)^2 .
\end{equation}
Here, $\mu_{u,v}$ is the weighted mean, while $\sigma_{u,v}$ is the weighted standard deviation of the local neighborhood $\Omega_{u,v}$. This normalization makes the defective-reference comparison less sensitive to local brightness and contrast variation, allowing the later stages to focus more on structural differences. Figure~\ref{fig:lcn_qualitative_examples} shows that LCN reduces lighting changes while preserving the main PCB structure.

\paragraph{Haar-band structural cues.}
To explicitly capture local PCB structure, a fixed Haar transform is applied to both normalized defective ($\hat{I}_d$) and reference ($\hat{I}_g$) images:
\begin{equation}
    B_d = \mathcal{H}(\hat{I}_d), \qquad
    B_g = \mathcal{H}(\hat{I}_g).
\end{equation}
This transform uses a grouped stride-2 convolution with four $2 \times 2$ filters per channel:

{\footnotesize
\begin{equation}
    K_{LL}=\frac{1}{2}
    \begin{bmatrix}
    1 & 1\\
    1 & 1
    \end{bmatrix}, \quad
    K_{LH}=\frac{1}{2}
    \begin{bmatrix}
    -1 & -1\\
    1 & 1
    \end{bmatrix}, \quad
    K_{HL}=\frac{1}{2}
    \begin{bmatrix}
    -1 & 1\\
    -1 & 1
    \end{bmatrix}, \quad
    K_{HH}=\frac{1}{2}
    \begin{bmatrix}
    1 & -1\\
    -1 & 1
    \end{bmatrix}
\end{equation}
}

\begin{figure}[H]
    \centering
    \includegraphics[width=0.35\linewidth]{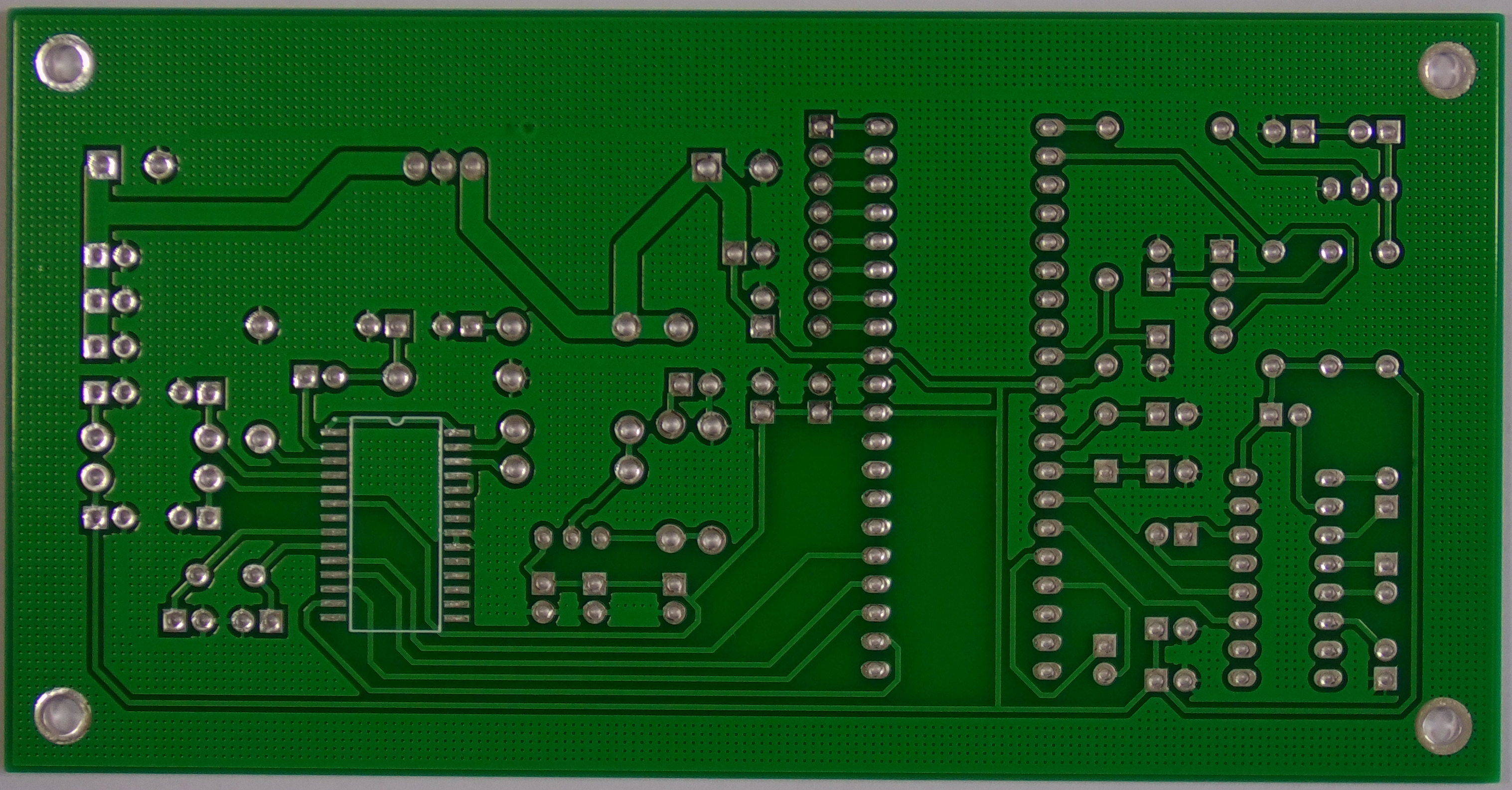}
    \caption{Example PCB image used to illustrate the Haar-band decomposition in Figure~\ref{fig:haar_example}.}
    \label{fig:haar_original_example}
\end{figure}

\begin{figure}[H]
    \centering
    \includegraphics[width=0.88\textwidth]{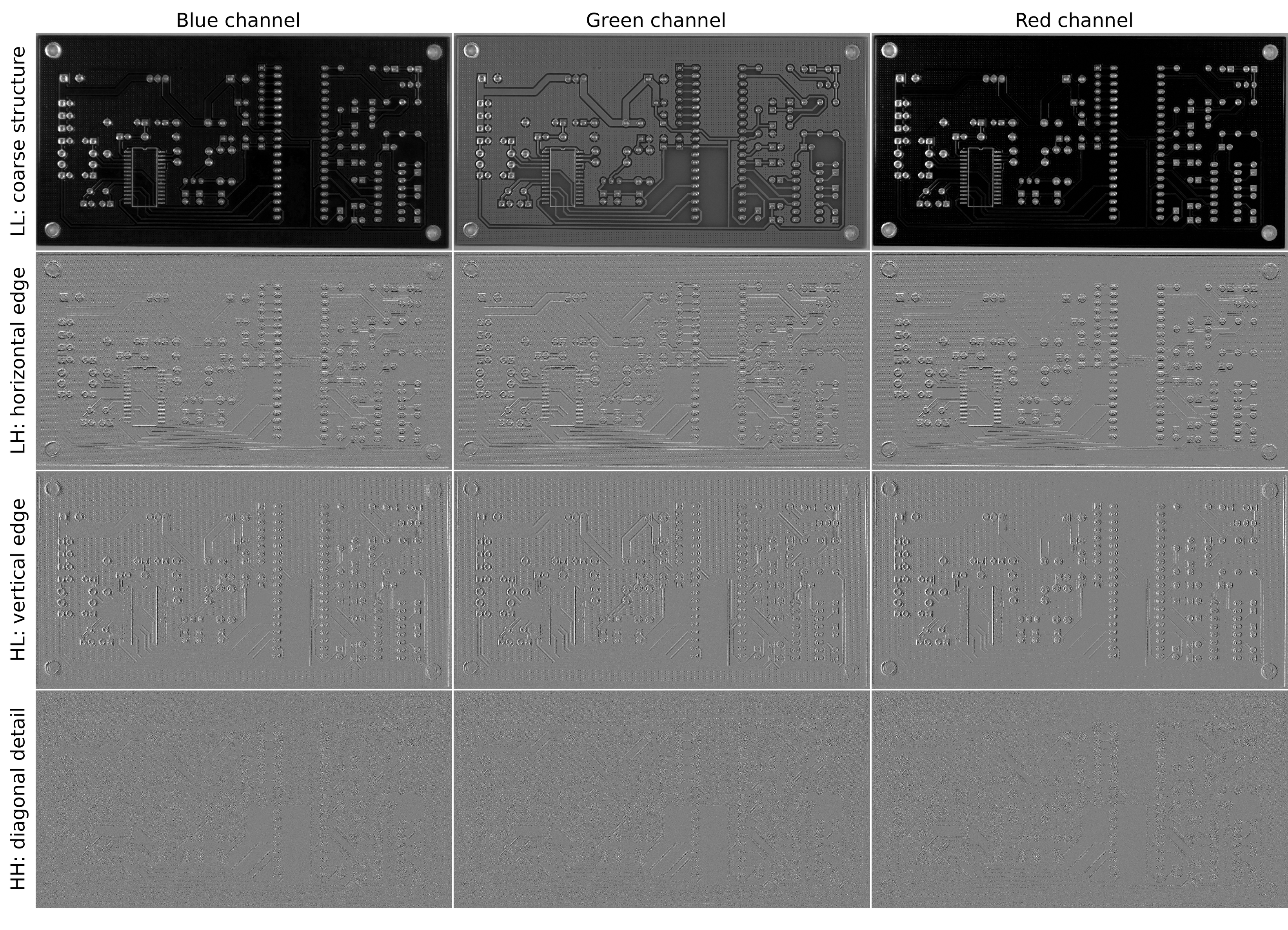}
    \caption{Haar-band decomposition of the PCB image in Figure~\ref{fig:haar_original_example}, shown as a $4\times3$ grid. The rows correspond to $LL$, $LH$, $HL$, and $HH$, while the columns correspond to the blue, green, and red channels.}
    \label{fig:haar_example}
\end{figure}

Applying these four filters to each non-overlapping $2 \times 2$ local region produces the $LL$, $LH$, $HL$, and $HH$ band outputs, respectively. As shown in Figure~\ref{fig:haar_example}, the $LL$ row preserves the broad layout and pad structures from the input PCB image in Figure~\ref{fig:haar_original_example}, while the $LH$, $HL$, and $HH$ rows capture horizontal, vertical, and diagonal changes in the PCB pattern. This helps RefDiffNet represent both the main PCB layout and small local variations that may indicate defects.

\begin{figure}[H]
\centering
\begin{tabular}{cc}
\bmvaHangBox{\fbox{\includegraphics[width=0.45\linewidth]{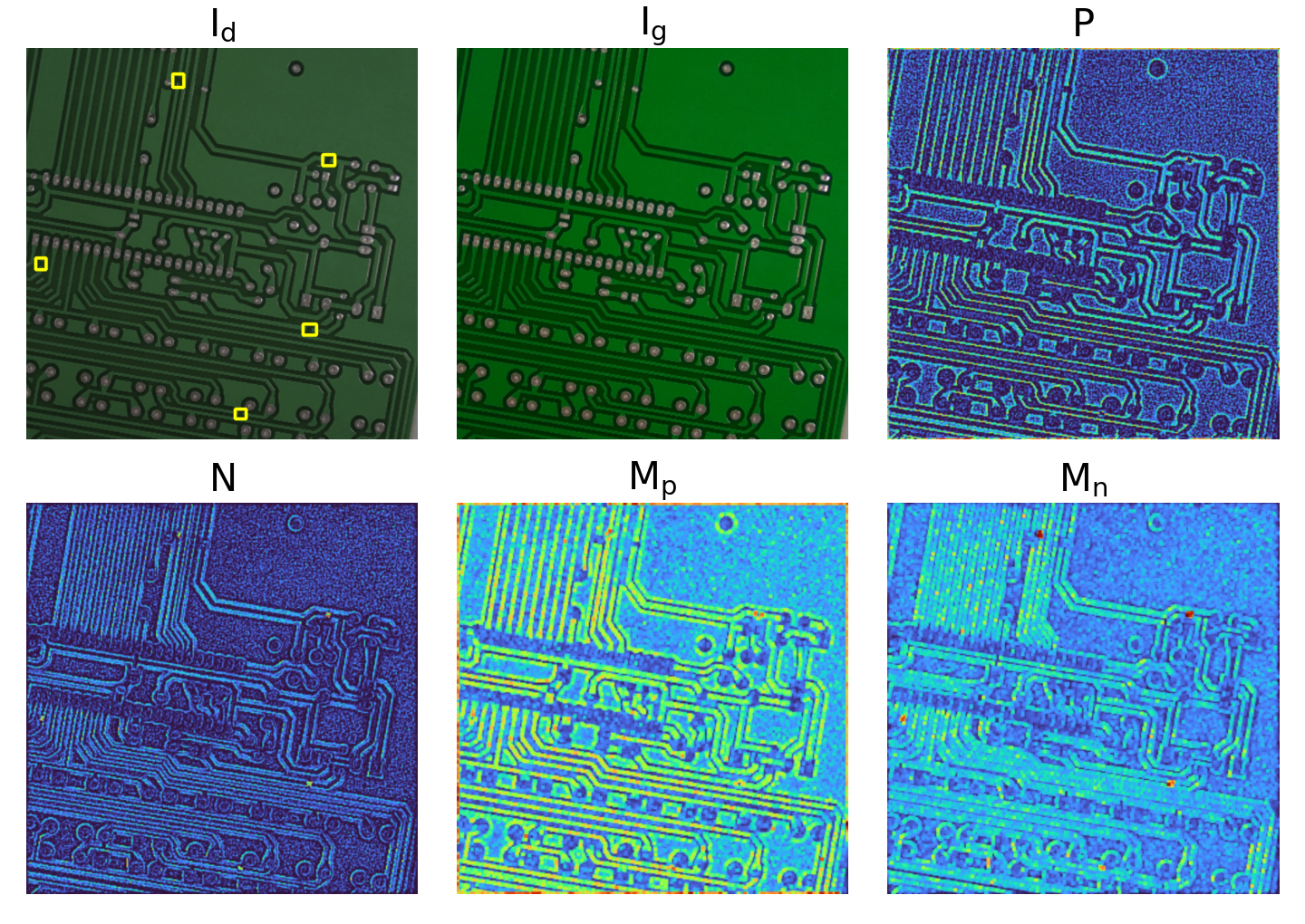}}} &
\bmvaHangBox{\fbox{\includegraphics[width=0.45\linewidth]{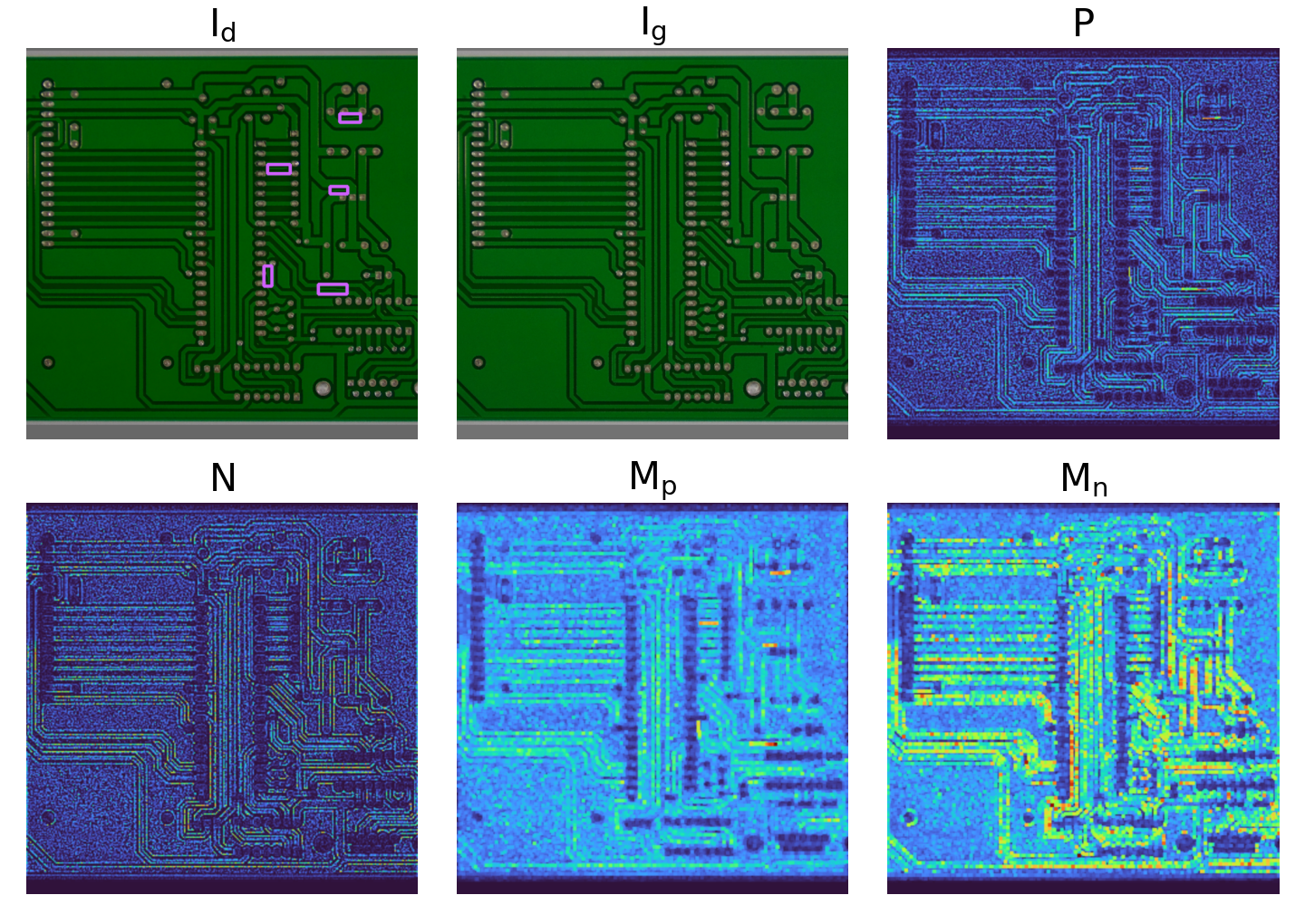}}} \\
(a) Open circuit & (b) Spurious copper
\end{tabular}
\caption{Qualitative visualization of signed residual and morphology-expanded cues.}
\label{fig:defect_cue_visualization}
\end{figure}

\paragraph{Signed residual cues.}
In addition to structural cues, RefDiffNet computes explicit defective-reference residuals at the same half-resolution scale as the Haar bands:
\begin{equation}
    I_d^{lr} = \mathrm{AvgPool}(\hat{I}_d), \qquad
    I_g^{lr} = \mathrm{AvgPool}(\hat{I}_g),
\end{equation}
\begin{equation}
    D^{lr} = I_d^{lr} - I_g^{lr}.
\end{equation}
Here, $I_d^{lr}$ and $I_g^{lr}$ denote the low-resolution, average-pooled versions of the LCN-normalized defective and reference images, whose difference gives the residual map $D^{lr}$. The residual is split into positive ($P$) and negative ($N$) components:
\begin{equation}
    P = \mathrm{ReLU}(D^{lr}), \qquad
    N = \mathrm{ReLU}(-D^{lr}).
\end{equation}
Keeping both positive and negative residual cues is important because defects appearing as missing material compared with the reference are more clearly captured by the negative cues, while defects appearing as extra material are more clearly captured by the positive cues. In Figure~\ref{fig:defect_cue_visualization}, the open-circuit defect in (a) is clearer in $N$, whereas the spurious-copper defect in (b) is clearer in $P$.

\paragraph{Morphology-expanded residual cues.}While $P$ and $N$ capture defective regions using positive and negative defective-reference differences, the highlighted defective regions in Figure~\ref{fig:defect_cue_visualization} show that these cues can occupy only a few pixels at the low-resolution scale. We therefore use $3 \times 3$ max-pooling with stride $1$ and padding $1$ to locally expand these cues and improve the visibility of defect regions without changing the spatial resolution.
\begin{equation}
    M_p = \mathrm{MaxPool}_{3 \times 3}(P), \qquad
    M_n = \mathrm{MaxPool}_{3 \times 3}(N).
    \label{eqn:maxpool}
\end{equation}

The final cue tensor $X \in \mathbb{R}^{12C \times \frac{H}{2} \times \frac{W}{2}}$ for RGB input is
\begin{equation}
    X = \mathrm{Concat}(B_d, B_g, P, N, M_p, M_n),
    \label{eq:cue_tensor}
\end{equation}
.

\subsection{Lightweight Encoder and Channel Recalibration}

The cue tensor $X$ is passed through a lightweight convolutional encoder, as shown in Figure~\ref{fig:encoder_spatial_gate_blocks}(a):
\begin{equation}
    F = \mathrm{Encoder}(X),
\end{equation}
where $F \in \mathbb{R}^{H_{\mathrm{hid}} \times \frac{H}{2} \times \frac{W}{2}}$ and $H_{\mathrm{hid}}=24$ in our implementation.

The encoder first applies a $1 \times 1$ pointwise convolution to project the $12C$ cue channels into $H_{\mathrm{hid}}$ channels, forming a compact representation of structural, signed residual, and morphology-expanded cues. Two depthwise-separable convolution blocks then refine this representation, where pointwise layers mix channel information and depthwise layers refine important spatial details with low computational cost~\cite{Howard2017MobileNets}. Since not all cue channels are equally useful for every defect type, RefDiffNet applies channel recalibration using a squeeze-and-excitation style channel gate~\cite{Hu2018SENet}, denoted as $\mathrm{SE}(\cdot)$, to reweight the encoded feature map.

\begin{equation}
    g_c = \mathrm{SE}(F), 
    \qquad
    g_c \in \mathbb{R}^{H_{\mathrm{hid}} \times 1 \times 1}.
\end{equation}
and the recalibrated feature map is
\begin{equation}
    F_c = F \odot g_c .
\end{equation}
This allows the model to emphasize useful structural or residual cue channels before predicting the enhancement maps.

\subsection{RGB Correction and Spatial Gate Prediction}

From the recalibrated feature map $F_c$, RefDiffNet predicts two complementary outputs. First, RefDiffNet predicts a low-resolution RGB correction map:
\begin{equation}
    \Delta^{lr} = \mathrm{RGBDelta}(F_c),
    \qquad
    \Delta^{lr} \in \mathbb{R}^{C \times \frac{H}{2} \times \frac{W}{2}}.
\end{equation}

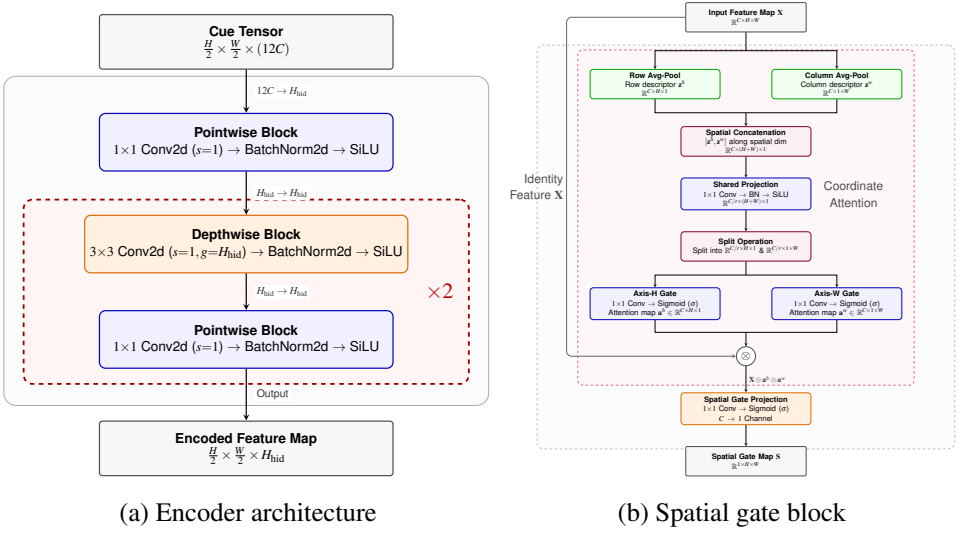
\begin{figure}[H]
    \centering

    \begin{minipage}[t]{0.50\textwidth}
    \centering
    \resizebox{\linewidth}{!}{%
    \begin{tikzpicture}[
        >=stealth,
        thick,
        node distance=0.85cm,
        font=\sffamily,
        tensor/.style={
            rectangle,
            draw=black!70,
            fill=gray!6,
            rounded corners=2pt,
            minimum height=3.6em,
            minimum width=6.8cm,
            align=center,
            font=\sffamily\small
        },
        pwconv/.style={
            rectangle,
            draw=blue!75!black,
            fill=blue!6,
            rounded corners=4pt,
            minimum height=3.8em,
            minimum width=6.8cm,
            align=center,
            font=\sffamily\small
        },
        dwconv/.style={
            rectangle,
            draw=orange!85!black,
            fill=orange!12,
            rounded corners=4pt,
            minimum height=3.8em,
            minimum width=6.8cm,
            align=center,
            font=\sffamily\small
        },
        edge_label/.style={
            right=5pt,
            font=\sffamily\scriptsize,
            text=black!80,
            fill=white,
            inner sep=1.5pt
        }
    ]

    \node[tensor] (input) {\textbf{Cue Tensor} \\ $\frac{H}{2}\times \frac{W}{2}\times (12C)$};

    \node[pwconv, below=1.0cm of input] (conv1) {\textbf{Pointwise Block} \\ $1{\times}1$ Conv2d ($s{=}1$) $\rightarrow$ BatchNorm2d $\rightarrow$ SiLU};

    \node[dwconv, below=1.0cm of conv1] (conv2) {\textbf{Depthwise Block} \\ $3{\times}3$ Conv2d ($s{=}1, g{=}H_{\mathrm{hid}}$) $\rightarrow$ BatchNorm2d $\rightarrow$ SiLU};
    \node[pwconv, below=0.85cm of conv2] (conv3) {\textbf{Pointwise Block} \\ $1{\times}1$ Conv2d ($s{=}1$) $\rightarrow$ BatchNorm2d $\rightarrow$ SiLU};

    \node[tensor, below=1.2cm of conv3] (output) {\textbf{Encoded Feature Map} \\ $\frac{H}{2}\times \frac{W}{2}\times H_{\mathrm{hid}}$};

    \draw[->] (input) -- (conv1) node[midway, edge_label] {$12C \rightarrow H_{\mathrm{hid}}$};
    \draw[->] (conv1) -- (conv2) node[midway, edge_label] {$H_{\mathrm{hid}} \rightarrow H_{\mathrm{hid}}$};
    \draw[->] (conv2) -- (conv3) node[midway, edge_label] {$H_{\mathrm{hid}} \rightarrow H_{\mathrm{hid}}$};
    \draw[->] (conv3) -- (output) node[midway, edge_label] {Output};

    \begin{scope}[on background layer]
        
        \node[
            rectangle,
            draw=gray!65,
            fill=gray!3,
            rounded corners=8pt,
            inner xsep=2.2cm,
            inner ysep=0.85cm,
            fit=(conv1) (conv3)
        ] (encoder_box) {};

        \node[
            rectangle,
            draw=red!60!black,
            dashed,
            very thick,
            fill=red!4,
            rounded corners=6pt,
            inner xsep=1.4cm,
            inner ysep=0.35cm,
            fit=(conv2) (conv3)
        ] (repeat_box) {};

        \node[
            anchor=south west,
            font=\sffamily\bfseries,
            text=black!75,
            yshift=2pt
        ] at (encoder_box.north west) {};

        \node[
            anchor=east,
            font=\sffamily\Large\bfseries,
            text=red!75!black
        ] at ([xshift=-0.25cm]repeat_box.east) {$\times 2$};

    \end{scope}

    \end{tikzpicture}
    }
    
    \vspace{0.5em}
    (a) Encoder architecture
    \end{minipage}
   \hspace{0.001\textwidth}
    \begin{minipage}[t]{0.46\textwidth}
    \centering
    \resizebox{\linewidth}{!}{%
    \begin{tikzpicture}[
        >=stealth,
        thick,
        font=\sffamily,
        tensor/.style={rectangle, draw=black!70, fill=gray!6, rounded corners=2pt, minimum height=3.5em, text width=4.8cm, align=center, font=\sffamily\small},
        pool/.style={rectangle, draw=green!65!black, fill=green!6, rounded corners=4pt, minimum height=3.5em, text width=5.2cm, align=center, font=\sffamily\small},
        op/.style={rectangle, draw=purple!70!black, fill=purple!6, rounded corners=4pt, minimum height=3.5em, text width=5.2cm, align=center, font=\sffamily\small},
        conv/.style={rectangle, draw=blue!75!black, fill=blue!6, rounded corners=4pt, minimum height=3.8em, text width=5.2cm, align=center, font=\sffamily\small},
        gate/.style={rectangle, draw=orange!85!black, fill=orange!12, rounded corners=4pt, minimum height=3.8em, text width=5.2cm, align=center, font=\sffamily\small},
        math/.style={circle, draw=black!70, fill=white, inner sep=1pt, minimum size=0.8cm, font=\bfseries\Large, align=center}
    ]

    \node[tensor] (input) at (0, 0) {\textbf{Input Feature Map $\mathbf{X}$} \\ $\mathbb{R}^{C \times H \times W}$};

    \node[pool] (pool_h) at (-3.8, -2.8) {\textbf{Row Avg-Pool} \\ Row descriptor $\mathbf{z}^h$ \\ $\mathbb{R}^{C \times H \times 1}$};

    \node[pool] (pool_w) at (3.8, -2.8) {\textbf{Column Avg-Pool} \\ Column descriptor $\mathbf{z}^w$ \\ $\mathbb{R}^{C \times 1 \times W}$};
    
    \node[op] (concat) at (0, -5.2) {\textbf{Spatial Concatenation} \\ $[\mathbf{z}^h, \mathbf{z}^w]$ along spatial dim \\ $\mathbb{R}^{C \times (H+W) \times 1}$};

    \node[conv] (shared) at (0, -7.4) {\textbf{Shared Projection} \\ $1{\times}1$ Conv $\rightarrow$ BN $\rightarrow$ SiLU \\ $\mathbb{R}^{C/r \times (H+W) \times 1}$};

    \node[op] (split) at (0, -9.6) {\textbf{Split Operation} \\ Split into $\mathbb{R}^{C/r \times H \times 1}$ \& $\mathbb{R}^{C/r \times 1 \times W}$};
    
    \node[conv] (conv_h) at (-3.8, -12.0) {\textbf{Axis-H Gate} \\ $1{\times}1$ Conv $\rightarrow$ Sigmoid ($\sigma$) \\ Attention map $\mathbf{a}^h \in \mathbb{R}^{C \times H \times 1}$};

    \node[conv] (conv_w) at (3.8, -12.0) {\textbf{Axis-W Gate} \\ $1{\times}1$ Conv $\rightarrow$ Sigmoid ($\sigma$) \\ Attention map $\mathbf{a}^w \in \mathbb{R}^{C \times 1 \times W}$};

    \node[math] (mult) at (0, -14.2) {$\otimes$};

    \node[gate] (final_conv) at (0, -16.4) {\textbf{Spatial Gate Projection} \\ $1{\times}1$ Conv $\rightarrow$ Sigmoid ($\sigma$) \\ $C \rightarrow 1$ Channel};

    \node[tensor] (output) at (0, -18.6) {\textbf{Spatial Gate Map $\mathbf{S}$} \\ $\mathbb{R}^{1 \times H \times W}$};


    \coordinate (in_split) at (0, -1.4);
    \draw[-] (input.south) -- (in_split);
    \draw[->] (in_split) -| (pool_h.north);
    \draw[->] (in_split) -| (pool_w.north);

    \coordinate (merge_concat) at (0, -4.0);
    \draw[-] (pool_h.south) |- (merge_concat);
    \draw[-] (pool_w.south) |- (merge_concat);
    \draw[->] (merge_concat) -- (concat.north);

    \draw[->] (concat.south) -- (shared.north);
    \draw[->] (shared.south) -- (split.north);

    \coordinate (out_split) at (0, -10.8);
    \draw[-] (split.south) -- (out_split);
    \draw[->] (out_split) -| (conv_h.north);
    \draw[->] (out_split) -| (conv_w.north);

    \coordinate (merge_mult) at (0, -13.2);
    \draw[-] (conv_h.south) |- (merge_mult);
    \draw[-] (conv_w.south) |- (merge_mult);
    \draw[->] (merge_mult) -- (mult.north);

    \draw[->, rounded corners=6pt, color=black!60, very thick]
        (input.west) -- (-7.5, 0) |- (mult.west)
        node[pos=0.25, left, align=right, font=\sffamily\Large, text=black!70] {Identity\\Feature $\mathbf{X}$};

    \draw[->] (mult.south) -- (final_conv.north) node[midway, right, font=\sffamily\small, text=black!80] {$\mathbf{X} \otimes \mathbf{a}^h \otimes \mathbf{a}^w$};
    \draw[->] (final_conv.south) -- (output.north);

    \begin{scope}[on background layer]
        
        \node[
            rectangle,
            draw=gray!50,
            dashed,
            fill=gray!2,
            rounded corners=8pt,
            inner xsep=2.2cm,
            inner ysep=1.0cm,
            fit=(pool_h) (pool_w) (final_conv)
        ] (outer_box) {};

        \node[
            rectangle,
            draw=purple!60,
            dashed,
            fill=purple!3,
            rounded corners=6pt,
            inner xsep=0.5cm,
            inner ysep=0.8cm,
            fit=(pool_h) (pool_w) (mult)
        ] (stem_box) {};

        \node[
            align=center,
            font=\sffamily\Large,
            text=black!70
        ] at (4.5, -7.4) {Coordinate\\[0.3ex]Attention};
        
    \end{scope}

    \end{tikzpicture}
    }
    
    \vspace{0.5em}
    (b) Spatial gate block
    \end{minipage}
    \vspace{0.5em}
    \caption{Encoder and spatial gate blocks used in RefDiffNet.}
    \label{fig:encoder_spatial_gate_blocks}
\end{figure}

This map represents a learned color-level correction for each spatial location. Intuitively, it tells the model how the defective image should be adjusted to make defect regions more visible. The RGB branch is defined as
\begin{equation}
    \mathrm{RGBDelta}(\cdot)
    =
    \tanh
    \left(
    \mathrm{Conv}_{1 \times 1}
    \left(
    \mathrm{ConvBNAct}(\cdot)
    \right)
    \right).
\end{equation}
Since the defective image is normalized to $[0,1]$, the correction is bounded to $[-1,1]$ by $\tanh$ and the enhanced image is clamped back to $[0,1]$. In parallel, RefDiffNet predicts a one-channel spatial gate:
\begin{equation}
    S^{lr} = \mathcal{G}(F_c),
    \qquad
    S^{lr} \in \mathbb{R}^{1 \times \frac{H}{2} \times \frac{W}{2}}.
\end{equation}
As shown in Figure~\ref{fig:encoder_spatial_gate_blocks}(b), the gate predictor $\mathcal{G}$ uses a coordinate-attention-based spatial gate before the final one-channel projection~\cite{hou2021coordinate}. This helps preserve row-wise and column-wise layout information, which is useful for PCB traces and copper regions. The spatial gate then controls the correction strength at each location, suppressing changes in normal PCB regions and emphasizing likely defect regions.

\subsection{Residual Enhancement and Detector Input}

The predicted RGB correction and spatial gate are upsampled to the original image resolution using bilinear interpolation:
\begin{equation}
    \Delta = \mathrm{Upsample}(\Delta^{lr}), \qquad
    S = \mathrm{Upsample}(S^{lr}),
\end{equation}
\begin{equation}
    \Delta \in \mathbb{R}^{C \times H \times W}, \qquad
    S \in \mathbb{R}^{1 \times H \times W}.
\end{equation}

The enhanced image is then obtained using the residual formulation in Eq.~\ref{eq:refdiffnet_output}. This design keeps the original defective image intact and adds only a learned, spatially controlled correction. The enhanced image $I_{\mathrm{out}}$ is finally passed to the downstream detector without modifying the detector architecture.

\begin{figure}[H]
    \centering
    \includegraphics[width=0.99\textwidth]{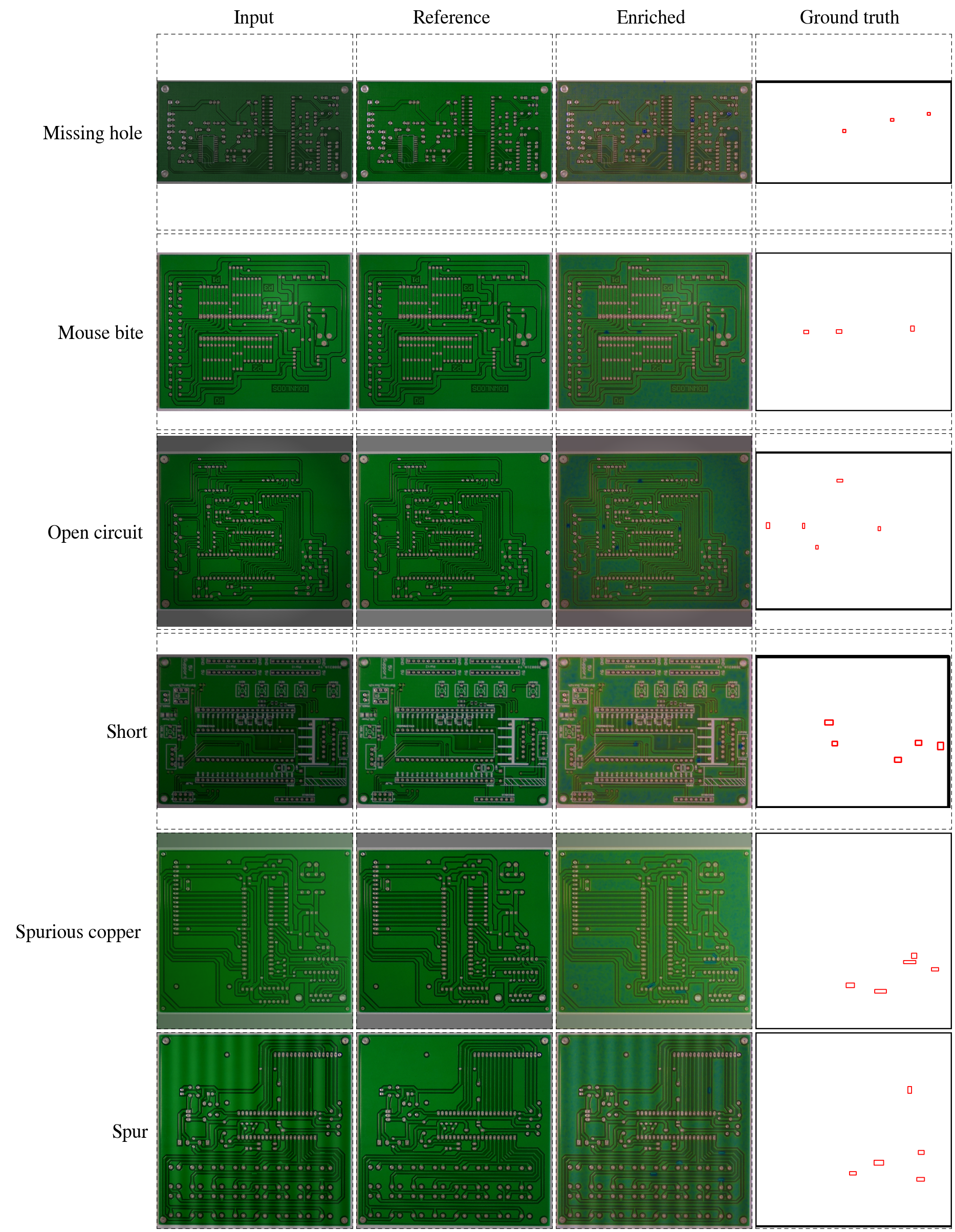}
    \caption{Qualitative visualization of RefDiffNet on representative PCB defect samples. Each row corresponds to one selected sample from a defect class, while the columns show, from left to right, the input defective image, the aligned reference image, the enriched output produced by RefDiffNet, and the ground-truth annotation.}
    \label{fig:qualitative_enhancement}
\end{figure}

\section{Results}
\subsection{Experimental setup}

We evaluate multiple detector families, including YOLOv8 to YOLOv26~\cite{yolo8,yolo9,yolo10,yolo11,yolo12,yolo26_ultralytics}, RT-DETR-L~\cite{zhao2024rtdetr}, and Faster R-CNN~\cite{fasterrcnn}. The YOLO variants and RT-DETR-L are trained using the Ultralytics framework~\cite{yolo26_ultralytics}, while Faster R-CNN uses the TorchVision \textit{Faster R-CNN ResNet-50 FPN} implementation~\cite{torchvision2016}. All detectors are initialized with COCO-pretrained weights and fine-tuned on the PCB datasets. For each detector, we train both the original model and the corresponding Original+RefDiffNet variant under the same training settings so that the effect of the proposed pre-backbone enhancement module can be measured directly. All models are trained for 200 epochs at $640 \times 640$ image resolution with batch size 8, using AdamW with an initial learning rate of 0.001, final learning-rate factor 0.01, weight decay $5\times10^{-4}$, cosine scheduling, 3 warmup epochs, and a fixed random seed of 42. For Original+RefDiffNet, the proposed module is inserted before the detector backbone: the defective image and its aligned reference image are passed to RefDiffNet, and the resulting enhanced image is then given to the detector. We report F1 score, mAP$_{50}$, and mAP$_{50:95}$, along with parameter count and FLOPs to measure computational overhead.

\begin{table}[H]
  \centering
  \caption{Baseline detectors vs. RefDiffNet-enhanced detectors on DeepPCB.}
  \scriptsize
  \setlength{\tabcolsep}{2.5pt}
  \renewcommand{\arraystretch}{1.12}
  \resizebox{\textwidth}{!}{%
  \begin{tabular}{lccccc|ccccc}
    \hline
    \textbf{\makecell[l]{Model}} &
    \multicolumn{5}{c|}{\textbf{Original}} &
    \multicolumn{5}{c}{\textbf{Original+RefDiffNet}} \\
    \cline{2-11}
    & \textbf{\makecell[c]{Param\\(M)}}
    & \textbf{\makecell[c]{FLOPs\\(G)}}
    & \textbf{\makecell[c]{F1\\(\%)}}
    & \textbf{\makecell[c]{mAP$_{50}$\\(\%)}}
    & \textbf{\makecell[c]{mAP$_{50:95}$\\(\%)}}
    & \textbf{\makecell[c]{Param\\(M)}}
    & \textbf{\makecell[c]{FLOPs\\(G)}}
    & \textbf{\makecell[c]{F1\\(\%)}}
    & \textbf{\makecell[c]{mAP$_{50}$\\(\%)}}
    & \textbf{\makecell[c]{mAP$_{50:95}$\\(\%)}} \\
    \hline

RT-DETR & 32.818 & 108.0 & 96.4 & 97.1 & 81.8
& 32.823 & 108.8
& \makecell[c]{96.7 {\footnotesize\textbf{[+0.3\%]}}}
& \makecell[c]{98.1 {\footnotesize\textbf{[+1.0\%]}}}
& \makecell[c]{82.7 {\footnotesize\textbf{[+1.1\%]}}} \\
\hline

Faster R-CNN & 41.325 & 181.5 & 94.4 & 95.3 & 76.5
& 41.329 & 182.2
& \makecell[c]{97.8 {\footnotesize\textbf{[+3.5\%]}}}
& \makecell[c]{97.8 {\footnotesize\textbf{[+2.6\%]}}}
& \makecell[c]{79.9 {\footnotesize\textbf{[+4.5\%]}}} \\
\hline

    YOLOv8n & 3.012 & 8.2 & 95.3 & 97.1 & 78.9 & 3.016 & 9.0 & \makecell[c]{98.3 {\footnotesize\textbf{[+3.1\%]}}} & \makecell[c]{99.2 {\footnotesize\textbf{[+2.2\%]}}} & \makecell[c]{82.8 {\footnotesize\textbf{[+4.9\%]}}} \\
YOLOv8s & 11.138 & 28.7 & 95.8 & 97.9 & 78.7 & 11.142 & 29.4 & \makecell[c]{98.6 {\footnotesize\textbf{[+2.9\%]}}} & \makecell[c]{99.1 {\footnotesize\textbf{[+1.2\%]}}} & \makecell[c]{81.5 {\footnotesize\textbf{[+3.6\%]}}} \\        YOLOv8m & 25.860 & 79.1 & 95.1 & 97.3 & 79.7 & 25.864 & 79.9 & \makecell[c]{98.7 {\footnotesize\textbf{[+3.8\%]}}} & \makecell[c]{99.2 {\footnotesize\textbf{[+2.0\%]}}} & \makecell[c]{81.4 {\footnotesize\textbf{[+2.1\%]}}} \\    YOLOv8l & 43.634 & 165.4 & 95.6 & 98.2 & 83.2 & 43.639 & 166.2 & \makecell[c]{98.4 {\footnotesize\textbf{[+2.9\%]}}} & \makecell[c]{99.1 {\footnotesize\textbf{[+0.9\%]}}} & \makecell[c]{83.8 {\footnotesize\textbf{[+0.7\%]}}} \\
            YOLOv8x & 68.158 & 258.1 & 96.6 & 98.4 & 81.0 & 68.163 & 258.9 & \makecell[c]{98.5 {\footnotesize\textbf{[+2.0\%]}}} & \makecell[c]{99.1 {\footnotesize\textbf{[+0.7\%]}}} & \makecell[c]{82.2 {\footnotesize\textbf{[+1.5\%]}}} \\
    \hline

    YOLOv9t & 2.007 & 7.9 & 95.7 & 97.6 & 82.4 & 2.011 & 8.6 & \makecell[c]{98.3 {\footnotesize\textbf{[+2.7\%]}}} & \makecell[c]{99.2 {\footnotesize\textbf{[+1.6\%]}}} & \makecell[c]{84.5 {\footnotesize\textbf{[+2.5\%]}}} \\
    YOLOv9s & 7.290 & 27.4 & 96.1 & 98.3 & 84.0 & 7.294 & 28.2 & \makecell[c]{98.6 {\footnotesize\textbf{[+2.6\%]}}} & \makecell[c]{99.0 {\footnotesize\textbf{[+0.7\%]}}} & \makecell[c]{85.6 {\footnotesize\textbf{[+1.9\%]}}} \\
    YOLOv9m & 20.163 & 77.6 & 94.7 & 97.9 & 80.0 & 20.167 & 78.3 & \makecell[c]{98.6 {\footnotesize\textbf{[+4.1\%]}}} & \makecell[c]{99.2 {\footnotesize\textbf{[+1.3\%]}}} & \makecell[c]{80.4 {\footnotesize\textbf{[+0.5\%]}}} \\
    YOLOv9c & 25.534 & 103.7 & 96.1 & 98.1 & 82.0 & 25.538 & 104.5 & \makecell[c]{98.9 {\footnotesize\textbf{[+2.9\%]}}} & \makecell[c]{99.3 {\footnotesize\textbf{[+1.2\%]}}} & \makecell[c]{84.6 {\footnotesize\textbf{[+3.2\%]}}} \\
    YOLOv9e & 58.150 & 192.7 & 95.7 & 98.0 & 84.0 & 58.154 & 193.5 & \makecell[c]{98.6 {\footnotesize\textbf{[+3.0\%]}}} & \makecell[c]{99.3 {\footnotesize\textbf{[+1.3\%]}}} & \makecell[c]{86.1 {\footnotesize\textbf{[+2.5\%]}}} \\
    \hline

    YOLOv10n & 2.709 & 8.4 & 95.1 & 97.6 & 81.6 & 2.714 & 9.2 & \makecell[c]{97.3 {\footnotesize\textbf{[+2.3\%]}}} & \makecell[c]{99.0 {\footnotesize\textbf{[+1.4\%]}}} & \makecell[c]{84.3 {\footnotesize\textbf{[+3.3\%]}}} \\
    YOLOv10s & 8.071 & 24.8 & 95.3 & 97.5 & 82.1 & 8.075 & 25.6 & \makecell[c]{98.2 {\footnotesize\textbf{[+3.0\%]}}} & \makecell[c]{99.1 {\footnotesize\textbf{[+1.6\%]}}} & \makecell[c]{84.5 {\footnotesize\textbf{[+2.9\%]}}} \\
    YOLOv10m & 16.491 & 64.0 & 95.9 & 98.2 & 82.1 & 16.495 & 64.8 & \makecell[c]{97.7 {\footnotesize\textbf{[+1.9\%]}}} & \makecell[c]{98.8 {\footnotesize\textbf{[+0.6\%]}}} & \makecell[c]{84.7 {\footnotesize\textbf{[+3.2\%]}}} \\
    YOLOv10b & 20.460 & 98.7 & 95.8 & 98.3 & 81.7 & 20.465 & 99.5 & \makecell[c]{98.4 {\footnotesize\textbf{[+2.7\%]}}} & \makecell[c]{99.2 {\footnotesize\textbf{[+0.9\%]}}} & \makecell[c]{84.6 {\footnotesize\textbf{[+3.5\%]}}} \\
    YOLOv10l & 25.775 & 127.2 & 96.3 & 98.6 & 82.0 & 25.779 & 128.0 & \makecell[c]{98.0 {\footnotesize\textbf{[+1.8\%]}}} & \makecell[c]{99.2 {\footnotesize\textbf{[+0.6\%]}}} & \makecell[c]{85.2 {\footnotesize\textbf{[+3.9\%]}}} \\
    YOLOv10x & 31.666 & 171.1 & 95.1 & 98.0 & 82.1 & 31.671 & 171.8 & \makecell[c]{98.1 {\footnotesize\textbf{[+3.2\%]}}} & \makecell[c]{99.0 {\footnotesize\textbf{[+1.0\%]}}} & \makecell[c]{85.8 {\footnotesize\textbf{[+4.5\%]}}} \\
    \hline

    YOLOv11n & 2.591 & 6.4 & 96.0 & 97.9 & 81.5 & 2.595 & 7.2 & \makecell[c]{98.5 {\footnotesize\textbf{[+2.6\%]}}} & \makecell[c]{99.2 {\footnotesize\textbf{[+1.3\%]}}} & \makecell[c]{84.2 {\footnotesize\textbf{[+3.3\%]}}} \\
    YOLOv11s & 9.430 & 21.6 & 95.6 & 98.2 & 81.6 & 9.434 & 22.3 & \makecell[c]{98.4 {\footnotesize\textbf{[+2.9\%]}}} & \makecell[c]{99.1 {\footnotesize\textbf{[+0.9\%]}}} & \makecell[c]{83.2 {\footnotesize\textbf{[+2.0\%]}}} \\
    YOLOv11m & 20.058 & 68.2 & 95.9 & 98.1 & 82.8 & 20.062 & 69.0 & \makecell[c]{98.3 {\footnotesize\textbf{[+2.5\%]}}} & \makecell[c]{99.2 {\footnotesize\textbf{[+1.1\%]}}} & \makecell[c]{84.1 {\footnotesize\textbf{[+1.6\%]}}} \\
    YOLOv11l & 25.315 & 87.3 & 96.6 & 98.5 & 83.3 & 25.319 & 88.1 & \makecell[c]{98.3 {\footnotesize\textbf{[+1.8\%]}}} & \makecell[c]{99.0 {\footnotesize\textbf{[+0.5\%]}}} & \makecell[c]{84.4 {\footnotesize\textbf{[+1.3\%]}}} \\
    YOLOv11x & 56.881 & 195.5 & 95.9 & 97.4 & 82.8 & 56.885 & 196.3 & \makecell[c]{98.2 {\footnotesize\textbf{[+2.4\%]}}} & \makecell[c]{98.9 {\footnotesize\textbf{[+1.5\%]}}} & \makecell[c]{84.2 {\footnotesize\textbf{[+1.7\%]}}} \\
    \hline

    YOLOv12n & 2.569 & 6.5 & 96.1 & 97.9 & 81.9 & 2.573 & 7.3 & \makecell[c]{98.0 {\footnotesize\textbf{[+2.0\%]}}} & \makecell[c]{98.9 {\footnotesize\textbf{[+1.0\%]}}} & \makecell[c]{82.6 {\footnotesize\textbf{[+0.9\%]}}} \\
    YOLOv12s & 9.255 & 21.5 & 96.0 & 98.1 & 82.2 & 9.260 & 22.3 & \makecell[c]{98.2 {\footnotesize\textbf{[+2.3\%]}}} & \makecell[c]{99.1 {\footnotesize\textbf{[+1.0\%]}}} & \makecell[c]{82.0 {\footnotesize\textbf{[-0.2\%]}}} \\
    YOLOv12m & 20.142 & 67.8 & 96.0 & 98.4 & 81.3 & 20.146 & 68.5 & \makecell[c]{97.8 {\footnotesize\textbf{[+1.9\%]}}} & \makecell[c]{99.0 {\footnotesize\textbf{[+0.6\%]}}} & \makecell[c]{83.1 {\footnotesize\textbf{[+2.2\%]}}} \\
    YOLOv12l & 26.394 & 89.4 & 95.4 & 98.1 & 81.0 & 26.398 & 90.2 & \makecell[c]{98.4 {\footnotesize\textbf{[+3.1\%]}}} & \makecell[c]{99.1 {\footnotesize\textbf{[+1.0\%]}}} & \makecell[c]{84.9 {\footnotesize\textbf{[+4.8\%]}}} \\
    YOLOv12x & 59.125 & 199.9 & 95.8 & 98.1 & 82.6 & 59.130 & 200.6 & \makecell[c]{98.4 {\footnotesize\textbf{[+2.7\%]}}} & \makecell[c]{99.3 {\footnotesize\textbf{[+1.2\%]}}} & \makecell[c]{84.3 {\footnotesize\textbf{[+2.1\%]}}} \\
    \hline

    YOLOv26n & 2.506 & 5.8 & 95.2 & 97.7 & 81.2 & 2.510 & 6.6 & \makecell[c]{97.1 {\footnotesize\textbf{[+2.0\%]}}} & \makecell[c]{98.8 {\footnotesize\textbf{[+1.1\%]}}} & \makecell[c]{83.6 {\footnotesize\textbf{[+3.0\%]}}} \\
    YOLOv26s & 9.953 & 22.5 & 96.1 & 98.0 & 80.8 & 9.957 & 23.3 & \makecell[c]{96.6 {\footnotesize\textbf{[+0.5\%]}}} & \makecell[c]{98.8 {\footnotesize\textbf{[+0.8\%]}}} & \makecell[c]{81.6 {\footnotesize\textbf{[+1.0\%]}}} \\
    YOLOv26m & 21.782 & 74.8 & 95.8 & 98.3 & 79.5 & 21.786 & 75.5 & \makecell[c]{97.6 {\footnotesize\textbf{[+1.9\%]}}} & \makecell[c]{99.0 {\footnotesize\textbf{[+0.7\%]}}} & \makecell[c]{80.8 {\footnotesize\textbf{[+1.6\%]}}} \\
    YOLOv26l & 26.186 & 93.2 & 96.2 & 98.6 & 82.7 & 26.190 & 93.9 & \makecell[c]{98.1 {\footnotesize\textbf{[+2.0\%]}}} & \makecell[c]{99.0 {\footnotesize\textbf{[+0.4\%]}}} & \makecell[c]{83.0 {\footnotesize\textbf{[+0.4\%]}}} \\
    YOLOv26x & 58.822 & 208.6 & 95.3 & 98.2 & 82.0 & 58.827 & 209.4 & \makecell[c]{98.0 {\footnotesize\textbf{[+2.8\%]}}} & \makecell[c]{99.0 {\footnotesize\textbf{[+0.8\%]}}} & \makecell[c]{85.0 {\footnotesize\textbf{[+3.7\%]}}} \\
    \hline
  \end{tabular}%
  }
  {
  \scriptsize
  \begin{minipage}{0.8\textwidth}
  \begin{flushleft}
  \textbf{Note:} Relative improvement is computed as $\frac{\text{Original+RefDiffNet} - \text{Original}}{\text{Original}} \times 100$. 
  \end{flushleft}
  \end{minipage}
  }
  \label{tab:deeppcb_original_refdiffnet}
\end{table}

\subsection{Datasets}
We evaluate RefDiffNet on two publicly available PCB defect detection datasets: DeepPCB~\cite{deeppcb} and HRIPCB~\cite{hripcb}. Both datasets provide paired reference and defective PCB images with bounding-box annotations, making them suitable for evaluating reference image-based defect detection. For both datasets, we use the original public data exactly as released, without adding, removing, resampling; therefore, the number of images remains unchanged.

\paragraph{DeepPCB.}
DeepPCB contains 1,500 aligned PCB image pairs. The original PCB images were captured using a linear-scan CCD at high resolution, then cropped into $640 \times 640$ patches and aligned through template matching. The dataset provides bounding-box annotations for six defect categories: open, short, mousebite, spur, pin hole, and spurious copper.

\paragraph{HRIPCB.}
HRIPCB is a public PCB defect dataset designed for detection, classification. The PCB images were captured using an industrial camera-based acquisition setup with controlled illumination, and defects were then created on the PCB images to build a labelled dataset. It contains 1,386 images covering six defect categories: missing hole, mouse bite, open circuit, short, spur, and spurious copper.

\begin{table}[H]
  \centering
  \caption{Baseline detectors vs. RefDiffNet-enhanced detectors on HRIPCB.}
  \scriptsize
  \setlength{\tabcolsep}{2.5pt}
  \renewcommand{\arraystretch}{1.12}
  \resizebox{\textwidth}{!}{%
  \begin{tabular}{lccccc|ccccc}
    \hline
    \textbf{\makecell[l]{Model}} &
    \multicolumn{5}{c|}{\textbf{Original}} &
    \multicolumn{5}{c}{\textbf{Original+RefDiffNet}} \\
    \cline{2-11}
    & \textbf{\makecell[c]{Param\\(M)}}
    & \textbf{\makecell[c]{FLOPs\\(G)}}
    & \textbf{\makecell[c]{F1\\(\%)}}
    & \textbf{\makecell[c]{mAP$_{50}$\\(\%)}}
    & \textbf{\makecell[c]{mAP$_{50:95}$\\(\%)}}
    & \textbf{\makecell[c]{Param\\(M)}}
    & \textbf{\makecell[c]{FLOPs\\(G)}}
    & \textbf{\makecell[c]{F1\\(\%)}}
    & \textbf{\makecell[c]{mAP$_{50}$\\(\%)}}
    & \textbf{\makecell[c]{mAP$_{50:95}$\\(\%)}} \\
    \hline

   RT-DETR & 32.818 & 108.0 & 90.2 & 92.8 & 52.8
    & 32.823 & 108.8
    & \makecell[c]{95.8 {\footnotesize\textbf{[+6.2\%]}}}
    & \makecell[c]{97.8 {\footnotesize\textbf{[+5.4\%]}}}
    & \makecell[c]{56.0 {\footnotesize\textbf{[+6.1\%]}}} \\
    \hline

    Faster R-CNN & 41.325 & 181.5 & 90.0 & 86.3 & 45.0
    & 41.329 & 182.2
    & \makecell[c]{94.6 {\footnotesize\textbf{[+5.1\%]}}}
    & \makecell[c]{94.8 {\footnotesize\textbf{[+9.9\%]}}}
    & \makecell[c]{51.0 {\footnotesize\textbf{[+13.2\%]}}} \\
    \hline
    
    YOLOv8n & 3.012 & 8.2 & 92.1 & 93.5 & 50.0 & 3.016 & 9.0 & \makecell[c]{95.1 {\footnotesize\textbf{[+3.3\%]}}} & \makecell[c]{97.2 {\footnotesize\textbf{[+4.0\%]}}} & \makecell[c]{56.7 {\footnotesize\textbf{[+13.4\%]}}} \\
    YOLOv8s & 11.138 & 28.7 & 94.9 & 96.7 & 55.1 & 11.142 & 29.4 & \makecell[c]{96.8 {\footnotesize\textbf{[+2.0\%]}}} & \makecell[c]{97.6 {\footnotesize\textbf{[+0.9\%]}}} & \makecell[c]{58.6 {\footnotesize\textbf{[+6.4\%]}}} \\
    YOLOv8m & 25.860 & 79.1 & 95.4 & 96.5 & 56.1 & 25.864 & 79.9 & \makecell[c]{97.3 {\footnotesize\textbf{[+2.0\%]}}} & \makecell[c]{97.7 {\footnotesize\textbf{[+1.2\%]}}} & \makecell[c]{59.0 {\footnotesize\textbf{[+5.2\%]}}} \\
    YOLOv8l & 43.634 & 165.4 & 96.4 & 97.2 & 57.3 & 43.639 & 166.2 & \makecell[c]{98.1 {\footnotesize\textbf{[+1.8\%]}}} & \makecell[c]{98.7 {\footnotesize\textbf{[+1.5\%]}}} & \makecell[c]{59.2 {\footnotesize\textbf{[+3.3\%]}}} \\
    YOLOv8x & 68.158 & 258.1 & 96.2 & 97.0 & 57.9 & 68.163 & 258.9 & \makecell[c]{97.8 {\footnotesize\textbf{[+1.7\%]}}} & \makecell[c]{98.3 {\footnotesize\textbf{[+1.3\%]}}} & \makecell[c]{60.4 {\footnotesize\textbf{[+4.3\%]}}} \\
    \hline

    YOLOv9t & 2.007 & 7.9 & 92.0 & 93.5 & 49.6 & 2.011 & 8.6 & \makecell[c]{96.8 {\footnotesize\textbf{[+5.2\%]}}} & \makecell[c]{97.2 {\footnotesize\textbf{[+4.0\%]}}} & \makecell[c]{57.3 {\footnotesize\textbf{[+15.5\%]}}} \\
    YOLOv9s & 7.290 & 27.4 & 96.2 & 96.8 & 54.4 & 7.294 & 28.2 & \makecell[c]{96.7 {\footnotesize\textbf{[+0.5\%]}}} & \makecell[c]{97.9 {\footnotesize\textbf{[+1.1\%]}}} & \makecell[c]{58.8 {\footnotesize\textbf{[+8.1\%]}}} \\
    YOLOv9m & 20.163 & 77.6 & 97.0 & 96.9 & 56.9 & 20.167 & 78.3 & \makecell[c]{98.1 {\footnotesize\textbf{[+1.1\%]}}} & \makecell[c]{98.4 {\footnotesize\textbf{[+1.5\%]}}} & \makecell[c]{59.8 {\footnotesize\textbf{[+5.1\%]}}} \\
    YOLOv9c & 25.534 & 103.7 & 96.8 & 97.3 & 57.2 & 25.538 & 104.5 & \makecell[c]{98.2 {\footnotesize\textbf{[+1.4\%]}}} & \makecell[c]{98.3 {\footnotesize\textbf{[+1.0\%]}}} & \makecell[c]{60.6 {\footnotesize\textbf{[+5.9\%]}}} \\
    YOLOv9e & 58.150 & 192.7 & 96.8 & 96.9 & 58.3 & 58.154 & 193.5 & \makecell[c]{98.4 {\footnotesize\textbf{[+1.7\%]}}} & \makecell[c]{98.1 {\footnotesize\textbf{[+1.2\%]}}} & \makecell[c]{59.9 {\footnotesize\textbf{[+2.7\%]}}} \\
    \hline

    YOLOv10n & 2.709 & 8.4 & 92.3 & 94.2 & 51.9 & 2.714 & 9.2 & \makecell[c]{95.2 {\footnotesize\textbf{[+3.1\%]}}} & \makecell[c]{97.4 {\footnotesize\textbf{[+3.4\%]}}} & \makecell[c]{57.5 {\footnotesize\textbf{[+10.8\%]}}} \\
    YOLOv10s & 8.071 & 24.8 & 95.7 & 95.9 & 55.3 & 8.075 & 25.6 & \makecell[c]{96.5 {\footnotesize\textbf{[+0.8\%]}}} & \makecell[c]{98.1 {\footnotesize\textbf{[+2.3\%]}}} & \makecell[c]{58.5 {\footnotesize\textbf{[+5.8\%]}}} \\
    YOLOv10m & 16.491 & 64.0 & 95.3 & 96.2 & 56.2 & 16.495 & 64.8 & \makecell[c]{97.2 {\footnotesize\textbf{[+2.0\%]}}} & \makecell[c]{98.3 {\footnotesize\textbf{[+2.2\%]}}} & \makecell[c]{58.5 {\footnotesize\textbf{[+4.1\%]}}} \\
    YOLOv10b & 20.460 & 98.7 & 95.6 & 97.2 & 56.1 & 20.465 & 99.5 & \makecell[c]{97.2 {\footnotesize\textbf{[+1.7\%]}}} & \makecell[c]{98.7 {\footnotesize\textbf{[+1.5\%]}}} & \makecell[c]{58.4 {\footnotesize\textbf{[+4.1\%]}}} \\
    YOLOv10l & 25.775 & 127.2 & 95.7 & 96.6 & 57.3 & 25.779 & 128.0 & \makecell[c]{96.9 {\footnotesize\textbf{[+1.3\%]}}} & \makecell[c]{98.2 {\footnotesize\textbf{[+1.7\%]}}} & \makecell[c]{59.8 {\footnotesize\textbf{[+4.4\%]}}} \\
    YOLOv10x & 31.666 & 171.1 & 96.7 & 96.9 & 57.2 & 31.671 & 171.8 & \makecell[c]{97.2 {\footnotesize\textbf{[+0.5\%]}}} & \makecell[c]{98.3 {\footnotesize\textbf{[+1.4\%]}}} & \makecell[c]{59.8 {\footnotesize\textbf{[+4.5\%]}}} \\
    \hline

    YOLOv11n & 2.591 & 6.4 & 93.2 & 94.5 & 51.4 & 2.595 & 7.2 & \makecell[c]{97.7 {\footnotesize\textbf{[+4.8\%]}}} & \makecell[c]{98.5 {\footnotesize\textbf{[+4.2\%]}}} & \makecell[c]{57.4 {\footnotesize\textbf{[+11.7\%]}}} \\
    YOLOv11s & 9.430 & 21.6 & 94.6 & 96.5 & 55.2 & 9.434 & 22.3 & \makecell[c]{97.2 {\footnotesize\textbf{[+2.7\%]}}} & \makecell[c]{98.7 {\footnotesize\textbf{[+2.3\%]}}} & \makecell[c]{59.7 {\footnotesize\textbf{[+8.2\%]}}} \\
    YOLOv11m & 20.058 & 68.2 & 96.1 & 96.9 & 56.7 & 20.062 & 69.0 & \makecell[c]{98.4 {\footnotesize\textbf{[+2.4\%]}}} & \makecell[c]{99.0 {\footnotesize\textbf{[+2.2\%]}}} & \makecell[c]{60.4 {\footnotesize\textbf{[+6.5\%]}}} \\
    YOLOv11l & 25.315 & 87.3 & 95.5 & 95.7 & 55.8 & 25.319 & 88.1 & \makecell[c]{97.5 {\footnotesize\textbf{[+2.1\%]}}} & \makecell[c]{98.6 {\footnotesize\textbf{[+3.0\%]}}} & \makecell[c]{60.1 {\footnotesize\textbf{[+7.7\%]}}} \\
    YOLOv11x & 56.881 & 195.5 & 95.6 & 96.2 & 57.3 & 56.885 & 196.3 & \makecell[c]{97.3 {\footnotesize\textbf{[+1.8\%]}}} & \makecell[c]{98.0 {\footnotesize\textbf{[+1.9\%]}}} & \makecell[c]{60.7 {\footnotesize\textbf{[+5.9\%]}}} \\
    \hline

    YOLOv12n & 2.569 & 6.5 & 94.4 & 95.1 & 51.8 & 2.573 & 7.3 & \makecell[c]{97.9 {\footnotesize\textbf{[+3.7\%]}}} & \makecell[c]{98.3 {\footnotesize\textbf{[+3.4\%]}}} & \makecell[c]{58.1 {\footnotesize\textbf{[+12.2\%]}}} \\
    YOLOv12s & 9.255 & 21.5 & 95.6 & 95.8 & 55.7 & 9.260 & 22.3 & \makecell[c]{97.4 {\footnotesize\textbf{[+1.9\%]}}} & \makecell[c]{97.5 {\footnotesize\textbf{[+1.8\%]}}} & \makecell[c]{59.9 {\footnotesize\textbf{[+7.5\%]}}} \\
    YOLOv12m & 20.142 & 67.8 & 96.6 & 97.1 & 57.4 & 20.146 & 68.5 & \makecell[c]{98.0 {\footnotesize\textbf{[+1.4\%]}}} & \makecell[c]{98.1 {\footnotesize\textbf{[+1.0\%]}}} & \makecell[c]{61.1 {\footnotesize\textbf{[+6.4\%]}}} \\
    YOLOv12l & 26.394 & 89.4 & 95.3 & 96.1 & 56.9 & 26.398 & 90.2 & \makecell[c]{98.4 {\footnotesize\textbf{[+3.3\%]}}} & \makecell[c]{98.6 {\footnotesize\textbf{[+2.6\%]}}} & \makecell[c]{60.1 {\footnotesize\textbf{[+5.6\%]}}} \\
    YOLOv12x & 59.125 & 199.9 & 95.7 & 96.6 & 58.1 & 59.130 & 200.6 & \makecell[c]{97.6 {\footnotesize\textbf{[+2.0\%]}}} & \makecell[c]{98.2 {\footnotesize\textbf{[+1.7\%]}}} & \makecell[c]{59.3 {\footnotesize\textbf{[+2.1\%]}}} \\
    \hline

    YOLOv26n & 2.506 & 5.8 & 89.3 & 91.3 & 48.9 & 2.510 & 6.6 & \makecell[c]{96.6 {\footnotesize\textbf{[+8.2\%]}}} & \makecell[c]{98.0 {\footnotesize\textbf{[+7.3\%]}}} & \makecell[c]{57.7 {\footnotesize\textbf{[+18.0\%]}}} \\
    YOLOv26s & 9.953 & 22.5 & 93.4 & 95.5 & 55.5 & 9.957 & 23.3 & \makecell[c]{97.7 {\footnotesize\textbf{[+4.6\%]}}} & \makecell[c]{98.5 {\footnotesize\textbf{[+3.1\%]}}} & \makecell[c]{58.8 {\footnotesize\textbf{[+5.9\%]}}} \\
    YOLOv26m & 21.782 & 74.8 & 96.0 & 97.7 & 56.6 & 21.786 & 75.5 & \makecell[c]{97.5 {\footnotesize\textbf{[+1.6\%]}}} & \makecell[c]{98.3 {\footnotesize\textbf{[+0.6\%]}}} & \makecell[c]{60.4 {\footnotesize\textbf{[+6.7\%]}}} \\
    YOLOv26l & 26.186 & 93.2 & 96.0 & 96.9 & 57.5 & 26.190 & 93.9 & \makecell[c]{97.5 {\footnotesize\textbf{[+1.6\%]}}} & \makecell[c]{98.1 {\footnotesize\textbf{[+1.2\%]}}} & \makecell[c]{59.7 {\footnotesize\textbf{[+3.8\%]}}} \\
    YOLOv26x & 58.822 & 208.6 & 97.0 & 97.1 & 57.7 & 58.827 & 209.4 & \makecell[c]{98.0 {\footnotesize\textbf{[+1.0\%]}}} & \makecell[c]{98.7 {\footnotesize\textbf{[+1.6\%]}}} & \makecell[c]{60.2 {\footnotesize\textbf{[+4.3\%]}}} \\
    \hline
  \end{tabular}%
  }
  {
  \scriptsize
  \begin{minipage}{0.8\textwidth}
  \begin{flushleft}
  \textbf{Note:} Relative improvement is computed as $\frac{\text{Original+RefDiffNet} - \text{Original}}{\text{Original}} \times 100$.
  \end{flushleft}
  \end{minipage}
  }
  \label{tab:hripcb_original_refdiffnet}
\end{table}

\subsection{Quantitative results}
\label{subsec:quantitative_results}

Tables~\ref{tab:deeppcb_original_refdiffnet} and~\ref{tab:hripcb_original_refdiffnet} show that RefDiffNet consistently improves detectors across model sizes and architectures, including YOLOv8--YOLOv26, RT-DETR, and Faster R-CNN. The added cost is minimal, increasing the detector by only $0.004$--$0.005$M parameters and $0.7$--$0.8$ GFLOPs, since RefDiffNet is used only as a lightweight pre-backbone module.

On DeepPCB, where the baseline detectors are already strong, RefDiffNet still improves F1, mAP$_{50}$, and mAP$_{50:95}$ across most detector families and model variants. YOLOv9e+Ref-\\DiffNet obtains the best DeepPCB result, with 98.6\% F1, 99.3\% mAP$_{50}$, and 86.1\% mAP$_{50:95}$. In addition to the YOLO variants, RT-DETR improves from 81.8\% to 82.7\% mAP$_{50:95}$, and Faster R-CNN improves from 76.5\% to 79.9\% mAP$_{50:95}$ after adding RefDiffNet. Notably, smaller RefDiffNet-enhanced models can outperform much larger original detectors; for example, YOLOv9s+RefDiffNet uses only 7.294M parameters and 28.2 GFLOPs, but achieves 85.6\% mAP$_{50:95}$, exceeding the original YOLOv26x with 58.822M parameters and 208.6 GFLOPs.

On HRIPCB, the gains are larger due to greater variation in board layout, defect appearance, and image orientation. The largest relative gains are observed for lightweight models, with mAP$_{50:95}$ improvements of 13.4\% on YOLOv8n, 15.5\% on YOLOv9t, 10.8\% on YOLOv10n, 11.7\% on YOLOv11n, 12.2\% on YOLOv12n, and 18.0\% on YOLOv26n. RT-DETR improves from 52.8\% to 56.0\% mAP$_{50:95}$, while Faster R-CNN improves from 45.0\% to 51.0\% mAP$_{50:95}$. The best HRIPCB mAP$_{50:95}$ is obtained by YOLOv12m+RefDiffNet with 61.1\%. This model uses only 20.146M parameters and 68.5 GFLOPs, yet outperforms much larger original detectors such as YOLOv8x, YOLOv12x, and YOLOv26x.

These results show that RefDiffNet improves detection beyond a single architecture or model scale. It also allows smaller and medium-sized models to match or surpass larger baseline detectors, while improving transformer-based and two-stage detectors as well, demonstrating that reference-guided input enhancement is a parameter-efficient and detector-agnostic way to improve PCB defect detection.

\section{Ablation Study}
All ablations are conducted on HRIPCB using YOLO26n. We train each variant for 100 epochs at $640 \times 640$ resolution with batch size 16. All detector settings are kept fixed; only the ablated RefDiffNet component is changed.

\noindent\textbf{Overall architecture ablation.}
We first ablate the main architectural components of RefDiffNet before introducing the full cue stack. This experiment uses only the defective image and the aligned golden reference image, while excluding Haar bands, signed residual cues, and morphology-expanded cues. Table~\ref{tab:ablation_enhancement_structure} shows the incremental construction of RefDiffNet by adding one component at a time. 

\begin{table}[H]
\centering
\caption{Overall RefDiffNet architecture ablation.}
\label{tab:ablation_enhancement_structure}
\renewcommand{\arraystretch}{1.18}
\setlength{\tabcolsep}{4pt}
\resizebox{\textwidth}{!}{
\begin{tabular}{l l c c c c c c c}
\toprule
\textbf{Architecture variant}
& \textbf{Output sent to detector}
& \textbf{Retain $I_d$}
& \textbf{LCN}
& \textbf{Enc.}
& \textbf{CG}
& \textbf{SG}
& \textbf{mAP$_{50}$}
& \textbf{mAP$_{50:95}$} \\
\midrule

Raw difference only
& $I_d - I_g$
& \xmark & \xmark & \xmark & \xmark & \xmark
& 0.519 & 0.218 \\

Defect image with raw difference
& $I_d + \alpha(I_d - I_g)$
& \cmark & \xmark & \xmark & \xmark & \xmark
& 0.817 & 0.411 \\

Defect image with LCN difference
& $I_d + \alpha(\hat{I}_d - \hat{I}_g)$
& \cmark & \cmark & \xmark & \xmark & \xmark
& 0.850 & 0.436 \\

Learnable encoder enhancement
& $I_d + \alpha R$
& \cmark & \cmark & \cmark & \xmark & \xmark
& 0.875 & 0.438 \\

Encoder with channel recalibration
& $I_d + \alpha R_{\mathrm{cg}}$
& \cmark & \cmark & \cmark & \cmark & \xmark
& 0.867 & 0.444 \\

Encoder with channel recalibration and spatial gate
& $I_d + \alpha G \odot R_{\mathrm{cg}}$
& \cmark & \cmark & \cmark & \cmark & \cmark
& \textbf{0.880} & 0.446 \\

Full architecture with RGB correction
& $I_d + \alpha G \odot R_{\mathrm{rgb}}$
& \cmark & \cmark & \cmark & \cmark & \cmark
& 0.873 & \textbf{0.449} \\

\bottomrule
\end{tabular}
}
\end{table}
Here, $I_d$ denotes the original defective image and $I_g$ denotes the aligned golden reference image. Directly sending only their raw difference, $I_d-I_g$, to the detector performs poorly, achieving only 0.218 mAP$_{50:95}$, because it removes much of the original PCB context. Therefore, the next variant keeps $I_d$ as the base image and adds a scaled defective-reference difference, where the learnable parameter $\alpha$ controls the strength of the added correction, improving mAP$_{50:95}$ to 0.411. This preserves the circuit structure while highlighting regions that deviate from the reference. Replacing the raw images with their locally contrast-normalized versions, $\hat{I}d$ and $\hat{I}g$, further improves mAP$_{50:95}$ to 0.436 because the residual becomes less sensitive to illumination and local contrast changes. The learnable encoder replaces fixed subtraction with a predicted residual enhancement, $R$, enabling the model to learn corrections that are more relevant for defect detection and increasing mAP$_{50:95}$ to 0.438.

Applying channel recalibration produces $R_{\mathrm{cg}}$, which emphasizes more useful feature channels and improves mAP$_{50:95}$ to 0.444, while the spatial gate $G$ selects where this correction should be applied through element-wise multiplication, denoted by $\odot$, further increasing mAP$_{50:95}$ to 0.446. Finally, the full model predicts an RGB correction map, $R_{\mathrm{rgb}}$, so the detector receives an enhanced image in the same format as the original defective image, achieving the best mAP$_{50:95}$ of 0.449.

\noindent\textbf{Feature cues.}
We next ablate the encoder input feature cues used by RefDiffNet while keeping the rest of the module fixed; only the encoder input tensors are changed. The full cue stack includes Haar-band features from the defective and reference images, $B_d$ and $B_g$, signed residual maps, $P$ and $N$, and their morphology-expanded maps, $M_p$ and $M_n$. Table~\ref{tab:cue_subset_ablation} shows that the full cue stack gives the best localization performance. Haar cues alone are effective for mAP$_{50}$, but combining structural cues with signed residuals and morphology-expanded maps yields the highest mAP$_{50:95}$, indicating more precise bounding-box localization.

\begin{table}[H]
\centering
\caption{Cue-subset ablation. }
\label{tab:cue_subset_ablation}
\renewcommand{\arraystretch}{1.16}
\setlength{\tabcolsep}{5pt}
\resizebox{\textwidth}{!}{
\begin{tabular}{l l c c c c c}
\toprule
\textbf{Variant}
& \textbf{Encoder input}
& \textbf{Haar}
& \textbf{Signed}
& \textbf{Morph.}
& \textbf{mAP$_{50}$}
& \textbf{mAP$_{50:95}$} \\
\midrule

Full cue stack
& $\mathrm{Concat}(B_d, B_g, P, N, M_p, M_n)$
& \cmark & \cmark & \cmark
& 0.887 & \textbf{0.461} \\

Remove Haar bands
& $\mathrm{Concat}(P, N, M_p, M_n)$
& \xmark & \cmark & \cmark
& 0.871 & 0.438 \\

Remove signed residual maps
& $\mathrm{Concat}(B_d, B_g, M_p, M_n)$
& \cmark & \xmark & \cmark
& 0.862 & 0.441 \\

Remove morphology maps
& $\mathrm{Concat}(B_d, B_g, P, N)$
& \cmark & \cmark & \xmark
& 0.874 & 0.444 \\

Only morphology maps
& $\mathrm{Concat}(M_p, M_n)$
& \xmark & \xmark & \cmark
& 0.853 & 0.436 \\

Only signed residual maps
& $\mathrm{Concat}(P, N)$
& \xmark & \cmark & \xmark
& 0.888 & 0.451 \\

Only Haar bands
& $\mathrm{Concat}(B_d, B_g)$
& \cmark & \xmark & \xmark
& \textbf{0.891} & 0.456 \\

\bottomrule
\end{tabular}
}
\end{table}

\noindent\textbf{Feature-stack design.} We next study whether RefDiffNet should receive separated cue channels or compact difference-based inputs. In the proposed design, the defective and reference Haar-band features are kept as separate inputs, $B_d$ and $B_g$, instead of being collapsed into $B_d-B_g$. Similarly, the residual cues are split into positive and negative branches, $(P,N)$, with separate morphology-expanded maps, $(M_p,M_n)$. 

Table~\ref{tab:cue_subset_ablation} shows that the full separated cue stack achieves the best mAP$_{50:95}$. Collapsing the Haar features into a single difference map reduces performance, suggesting that $B_d$ and $B_g$ contain complementary structural information that is useful when kept separate. In addition, neither Haar differences nor signed residual maps are sufficient on their own. The best result is obtained when the encoder receives the full separated cue stack, $\mathrm{Concat}(B_d, B_g, P, N, M_p, M_n)$.

\begin{table}[H]
\centering
\caption{Feature-stack redesign ablation.}
\label{tab:feature_stack_redesign}
\renewcommand{\arraystretch}{1.18}
\setlength{\tabcolsep}{4pt}
\resizebox{\textwidth}{!}{
\begin{tabular}{l l c c}
\toprule
\textbf{Variant}
& \textbf{Encoder input}
& \textbf{mAP$_{50}$}
& \textbf{mAP$_{50:95}$} \\
\midrule

Full separated cue stack
& $\mathrm{Concat}(B_d, B_g, P, N, M_p, M_n)$
& \textbf{0.887} & \textbf{0.461} \\

Haar difference with positive branch
& $\mathrm{Concat}(B_d-B_g, P, M_p)$
& 0.828 & 0.414 \\

Haar difference with signed residual
& $\mathrm{Concat}(B_d-B_g, D^{lr}, M_p)$
& 0.861 & 0.441 \\

Reversed Haar difference with negative branch
& $\mathrm{Concat}(B_g-B_d, N, M_n)$
& 0.854 & 0.438 \\

Reversed Haar difference with signed negative residual
& $\mathrm{Concat}(B_g-B_d, -D^{lr}, M_n)$
& 0.832 & 0.421 \\

Separated Haar bands with signed residuals
& $\mathrm{Concat}(B_d, B_g, D^{lr}, -D^{lr}, \widetilde{M}_p, \widetilde{M}_n)$
& 0.878 & 0.444 \\

\bottomrule
\end{tabular}
}
\end{table}

\noindent\textbf{Spatial-gate attention.}
Finally, we evaluate the attention mechanism used to generate the spatial enhancement gate. The feature stack and encoder are fixed, and only the spatial-gate attention module is changed.
\begin{table}[H]
\centering
\caption{Ablation of the spatial-gate attention mechanism.}
\label{tab:attention_mechanism_ablation}
\renewcommand{\arraystretch}{1.18}
\setlength{\tabcolsep}{4pt}
\resizebox{0.5\textwidth}{!}{
\begin{tabular}{l c c}
\toprule
\textbf{Attention mechanism}
& \textbf{mAP$_{50}$}
& \textbf{mAP$_{50:95}$} \\
\midrule

Convolutional Block Attention Module~\cite{woo2018cbam}
& 0.831 & 0.450 \\

Efficient Channel Attention~\cite{wang2020eca}
& 0.793 & 0.431 \\

Squeeze-and-Excitation Attention~\cite{Hu2018SENet}
& 0.808 & 0.427 \\

Selective Kernel Attention~\cite{li2019selective}
& 0.819 & 0.442 \\

Coordinate Attention~\cite{hou2021coordinate}
& \textbf{0.887} & \textbf{0.461} \\

\bottomrule
\end{tabular}
}
\end{table}

Table~\ref{tab:attention_mechanism_ablation} shows that coordinate attention obtains the best mAP$_{50:95}$.
This is likely because coordinate attention, illustrated in Fig.~\ref{fig:encoder_spatial_gate_blocks}(b), encodes positional information separately along the height and width directions instead of collapsing all spatial information into a single global descriptor. This helps the spatial gate distinguish small defect regions from surrounding circuit patterns more precisely.

\section{Conclusion}
\label{sec:conclusion}

We presented \textbf{RefDiffNet}, a lightweight, plug-and-play pre-backbone enhancement module for reference image-guided PCB defect detection. By using the reference image to expose defect regions before passing the enhanced image to an off-the-shelf detector, RefDiffNet bridges classical template-based inspection and modern deep learning without altering the downstream detector architecture. RefDiffNet extracts structural, residual, and morphology-expanded cues from the defective-reference image pair, fuses them through a lightweight gated encoder, and produces an RGB correction map that is selectively applied through a coordinate-attention spatial gate. The entire module introduces only $0.004$--$0.005$M additional parameters and $0.7$--$0.8$ GFLOPs, making the overhead negligible relative to the evaluated detectors.

Across HRIPCB and DeepPCB, RefDiffNet improves performance across detector families, including one-stage YOLOv8--YOLOv26 detectors, transformer-based RT-DETR, and two-stage Faster R-CNN. Relative mAP$_{50:95}$ gains reach up to 18\% on HRIPCB, with smaller enhanced models frequently surpassing much larger unenhanced baselines. Ablations confirm that each component, including LCN, the gated encoder, the full cue stack, and coordinate-attention spatial gating, contributes meaningfully to the final performance. Overall, RefDiffNet demonstrates that reference-guided input enhancement is a simple, efficient, and detector-agnostic strategy for improving PCB defect detection, complementary to advances in detector architecture.

\bibliography{egbib}

@ARTICLE{Ling2023Survey,
  author={Ling, Qin and Isa, Nor Ashidi Mat},
  journal={IEEE Access}, 
  title={Printed Circuit Board Defect Detection Methods Based on Image Processing, Machine Learning and Deep Learning: A Survey}, 
  year={2023},
  volume={11},
  number={},
  pages={15921-15944},
  doi={10.1109/ACCESS.2023.3245093}}

@ARTICLE{chensurvey,
  author={Chen, Xing and Wu, Yonglei and He, Xingyou and Ming, Wuyi},
  journal={IEEE Access}, 
  title={A Comprehensive Review of Deep Learning-Based PCB Defect Detection}, 
  year={2023},
  volume={11},
  number={},
  pages={139017-139038},
  doi={10.1109/ACCESS.2023.3339561}}

@misc{deeppcb,
      title={Online PCB Defect Detector On A New PCB Defect Dataset}, 
      author={Sanli Tang and Fan He and Xiaolin Huang and Jie Yang},
      year={2019},
      eprint={1902.06197},
      archivePrefix={arXiv},
      primaryClass={cs.CV},
      url={https://arxiv.org/abs/1902.06197}, 
}

@article{hripcb,
author = {Huang, Weibo and Wei, Peng and Zhang, Manhua and Liu, Hong},
title = {HRIPCB: a challenging dataset for PCB defects detection and classification},
journal = {The Journal of Engineering},
volume = {2020},
number = {13},
pages = {303-309},
doi = {https://doi.org/10.1049/joe.2019.1183},
url = {https://ietresearch.onlinelibrary.wiley.com/doi/abs/10.1049/joe.2019.1183},
eprint = {https://ietresearch.onlinelibrary.wiley.com/doi/pdf/10.1049/joe.2019.1183},
year = {2020}
}

@misc{changechip,
      title={ChangeChip: A Reference-Based Unsupervised Change Detection for PCB Defect Detection}, 
      author={Yehonatan Fridman and Matan Rusanovsky and Gal Oren},
      year={2021},
      eprint={2109.05746},
      archivePrefix={arXiv},
      primaryClass={cs.CV},
      url={https://arxiv.org/abs/2109.05746}, 
}

@misc{fasterrcnn,
      title={Faster R-CNN: Towards Real-Time Object Detection with Region Proposal Networks}, 
      author={Shaoqing Ren and Kaiming He and Ross Girshick and Jian Sun},
      year={2016},
      eprint={1506.01497},
      archivePrefix={arXiv},
      primaryClass={cs.CV},
      url={https://arxiv.org/abs/1506.01497}, 
}

@software{yolo26_ultralytics,
  author = {Glenn Jocher and Jing Qiu},
  title = {Ultralytics YOLO26},
  version = {26.0.0},
  year = {2026},
  url = {https://github.com/ultralytics/ultralytics},
  orcid = {0000-0001-5950-6979, 0000-0003-3783-7069},
  license = {AGPL-3.0}
}

@INPROCEEDINGS{yolo8,
  author={Varghese, Rejin and M., Sambath},
  booktitle={2024 International Conference on Advances in Data Engineering and Intelligent Computing Systems (ADICS)}, 
  title={YOLOv8: A Novel Object Detection Algorithm with Enhanced Performance and Robustness}, 
  year={2024},
  volume={},
  number={},
  pages={1-6},
  doi={10.1109/ADICS58448.2024.10533619}}

@misc{yolo9,
      title={YOLOv9: Learning What You Want to Learn Using Programmable Gradient Information}, 
      author={Chien-Yao Wang and I-Hau Yeh and Hong-Yuan Mark Liao},
      year={2024},
      eprint={2402.13616},
      archivePrefix={arXiv},
      primaryClass={cs.CV},
      url={https://arxiv.org/abs/2402.13616}, 
}

@misc{yolo10,
      title={YOLOv10: Real-Time End-to-End Object Detection}, 
      author={Ao Wang and Hui Chen and Lihao Liu and Kai Chen and Zijia Lin and Jungong Han and Guiguang Ding},
      year={2024},
      eprint={2405.14458},
      archivePrefix={arXiv},
      primaryClass={cs.CV},
      url={https://arxiv.org/abs/2405.14458}, 
}

@misc{yolo11,
      title={YOLOv11: An Overview of the Key Architectural Enhancements}, 
      author={Rahima Khanam and Muhammad Hussain},
      year={2024},
      eprint={2410.17725},
      archivePrefix={arXiv},
      primaryClass={cs.CV},
      url={https://arxiv.org/abs/2410.17725}, 
}

@misc{yolo12,
      title={YOLOv12: Attention-Centric Real-Time Object Detectors}, 
      author={Yunjie Tian and Qixiang Ye and David Doermann},
      year={2025},
      eprint={2502.12524},
      archivePrefix={arXiv},
      primaryClass={cs.CV},
      url={https://arxiv.org/abs/2502.12524}, 
}

@article{cdiyolo,
author = {Xiao, Gaoshang and Hou, Shuling and Zhou, Huiying},
year = {2024},
month = {03},
pages = {},
title = {PCB defect detection algorithm based on CDI-YOLO},
volume = {14},
journal = {Scientific Reports},
doi = {10.1038/s41598-024-57491-3}
}

@article{yolobfrv,
author = {Liu, Jiaxin and Kang, Bingyu and Liu, Chao and Peng, Xunhui and Bai, Yan},
year = {2024},
month = {09},
pages = {6055},
title = {YOLO-BFRV: An Efficient Model for Detecting Printed Circuit Board Defects},
volume = {24},
journal = {Sensors},
doi = {10.3390/s24186055}
}

@Article{improvedyolo12,
AUTHOR = {Chen, Zhi and Liu, Bingxiang},
TITLE = {A High-Accuracy PCB Defect Detection Algorithm Based on Improved YOLOv12},
JOURNAL = {Symmetry},
VOLUME = {17},
YEAR = {2025},
NUMBER = {7},
ARTICLE-NUMBER = {978},
URL = {https://www.mdpi.com/2073-8994/17/7/978},
ISSN = {2073-8994},
DOI = {10.3390/sym17070978}
}

@article{He2025Review,
author = {He, Zihan and Lian, Yudong and Wang, Yulei and Lu, Zhiwei},
year = {2025},
month = {07},
pages = {106437},
title = {A Comprehensive Review of Research on Surface Defect Detection of PCBs Based on Machine Vision},
volume = {27},
journal = {Results in Engineering},
doi = {10.1016/j.rineng.2025.106437}
}

@article{Chauhan2011Subtraction,
author = {Pal, Ajay and Chauhan, Singh and Bhardwaj, Sharat},
year = {2011},
month = {07},
pages = {},
title = {Detection of Bare PCB Defects by Image Subtraction Method using Machine Vision},
volume = {2},
journal = {Proceedings of the World Congress on Engineering 2011, WCE 2011}
}

@INPROCEEDINGS{Putera2010Morphology,
  author={Indera Putera, S.H and Ibrahim, Z.},
  booktitle={2010 2nd International Conference on Education Technology and Computer}, 
  title={Printed circuit board defect detection using mathematical morphology and MATLAB image processing tools}, 
  year={2010},
  volume={5},
  number={},
  pages={V5-359-V5-363},
  doi={10.1109/ICETC.2010.5530052}}

@article{Ibrahim2005Wavelet,
author = {Ibrahim, Zuwairie and Al-Attas, Syed},
year = {2005},
month = {04},
pages = {201-213},
title = {Wavelet-based printed circuit board inspection algorithm},
volume = {12},
journal = {Integrated Computer-Aided Engineering},
doi = {10.3233/ICA-2005-12206}
}

@article{Deng2018AutoVRS,
  title={Building an Automatic Defect Verification System Using Deep Neural Network for PCB Defect Classification},
  author={Yu-Shan Deng and An-Chun Luo and M. J. Dai},
  journal={2018 4th International Conference on Frontiers of Signal Processing (ICFSP)},
  year={2018},
  pages={145-149},
  url={https://api.semanticscholar.org/CorpusID:54215616}
}

@article{Miao2021CSSNet,
author = {Miao, Yilin and Liu, Zhewei and Wu, Xiangning and Gao, Jie},
title = {Cost-Sensitive Siamese Network for PCB Defect Classification},
journal = {Computational Intelligence and Neuroscience},
volume = {2021},
number = {1},
pages = {7550670},
doi = {https://doi.org/10.1155/2021/7550670},
url = {https://onlinelibrary.wiley.com/doi/abs/10.1155/2021/7550670},
eprint = {https://onlinelibrary.wiley.com/doi/pdf/10.1155/2021/7550670},
year = {2021}
}

@article{Ding2020,
author = {Ding, Runwei and Zhang, Can and Zhu, Qisheng and Liu, Hong},
title = {Unknown defect detection for printed circuit board based on multi-scale deep similarity measure method},
journal = {The Journal of Engineering},
volume = {2020},
number = {13},
pages = {388-393},
doi = {https://doi.org/10.1049/joe.2019.1188},
url = {https://ietresearch.onlinelibrary.wiley.com/doi/abs/10.1049/joe.2019.1188},
eprint = {https://ietresearch.onlinelibrary.wiley.com/doi/pdf/10.1049/joe.2019.1188},
year = {2020}
}

@misc{DVQI2023,
      title={DVQI: A Multi-task, Hardware-integrated Artificial Intelligence System for Automated Visual Inspection in Electronics Manufacturing}, 
      author={Audrey Chung and Francis Li and Jeremy Ward and Andrew Hryniowski and Alexander Wong},
      year={2023},
      eprint={2312.09232},
      archivePrefix={arXiv},
      primaryClass={cs.CV},
      url={https://arxiv.org/abs/2312.09232}, 
}

@ARTICLE{DeepSiameseWelding2022,
  author={Ling, Zhigang and Zhang, Aoran and Ma, Dexin and Shi, Yuxin and Wen, He},
  journal={IEEE Transactions on Instrumentation and Measurement}, 
  title={Deep Siamese Semantic Segmentation Network for PCB Welding Defect Detection}, 
  year={2022},
  volume={71},
  number={},
  pages={1-11},
  doi={10.1109/TIM.2022.3154814}}

@article{Lim2023DeepContext,
  author  = {Lim, JiaYou and Lim, JunYi and Baskaran, Vishnu Monn and Wang, Xin},
  title   = {A Deep Context Learning Based PCB Defect Detection Model with Anomalous Trend Alarming System},
  journal = {Results in Engineering},
  volume  = {17},
  pages   = {100968},
  year    = {2023},
  doi     = {10.1016/j.rineng.2023.100968}
}

@article{Feng2023DDTR,
  author  = {Feng, Bo and Cai, Jian},
  title   = {PCB Defect Detection via Local Detail and Global Dependency Information},
  journal = {Sensors},
  volume  = {23},
  number  = {18},
  pages   = {7755},
  year    = {2023},
  doi     = {10.3390/s23187755}
}

@article{Yi2024YOLOv8DEE,
  author  = {Yi, Feifan and Mohamed, Ahmad Sufril Azlan and Noor, Mohd Halim Mohd and Ani, Fakhrozi Che and Zolkefli, Zol Effendi},
  title   = {YOLOv8-DEE: A High-Precision Model for Printed Circuit Board Defect Detection},
  journal = {PeerJ Computer Science},
  volume  = {10},
  pages   = {e2548},
  year    = {2024},
  doi     = {10.7717/peerj-cs.2548}
}

@article{Hou2025EffNetPCB,
  author  = {Hou, Yingqiang and Zhang, Xindong},
  title   = {A Lightweight and High-Accuracy Framework for Printed Circuit Board Defect Detection},
  journal = {Engineering Applications of Artificial Intelligence},
  volume  = {148},
  pages   = {110375},
  year    = {2025},
  doi     = {10.1016/j.engappai.2025.110375}
}

@article{Luo2025LiteDETR,
  author  = {Luo, Tao and Zhou, Yongbing and Shi, Donglin and Yun, Qinglin and Wang, Shuying and Zhang, Jian and Ding, Guofu},
  title   = {A Lightweight Defect Detection Transformer for Printed Circuit Boards Combining Image Feature Augmentation and Refined Cross-Scale Feature Fusion},
  journal = {Engineering Applications of Artificial Intelligence},
  pages   = {111128},
  year    = {2025},
  doi     = {10.1016/j.engappai.2025.111128}
}

@article{Zhang2025MSADETR,
  author  = {Zhang, Renjie and Gong, Yanjue and Zhao, Fu and Fan, Jinkai},
  title   = {MSA-DETR: Multi-Scale Alignment Detection Transformer for PCB Defect Detection with Enhanced Cross-Scale Feature Fusion},
  journal = {Measurement Science and Technology},
  volume  = {36},
  pages   = {125401},
  year    = {2025},
  doi     = {10.1088/1361-6501/ae2152}
}

@INPROCEEDINGS{lcn,
  author={Jarrett, Kevin and Kavukcuoglu, Koray and Ranzato, Marc'Aurelio and LeCun, Yann},
  booktitle={2009 IEEE 12th International Conference on Computer Vision}, 
  title={What is the best multi-stage architecture for object recognition?}, 
  year={2009},
  volume={},
  number={},
  pages={2146-2153},
  doi={10.1109/ICCV.2009.5459469}}

@misc{Howard2017MobileNets,
      title={MobileNets: Efficient Convolutional Neural Networks for Mobile Vision Applications}, 
      author={Andrew G. Howard and Menglong Zhu and Bo Chen and Dmitry Kalenichenko and Weijun Wang and Tobias Weyand and Marco Andreetto and Hartwig Adam},
      year={2017},
      eprint={1704.04861},
      archivePrefix={arXiv},
      primaryClass={cs.CV},
      url={https://arxiv.org/abs/1704.04861}, 
}

@inproceedings{Hu2018SENet,
  author    = {Hu, Jie and Shen, Li and Sun, Gang},
  title     = {Squeeze-and-Excitation Networks},
  booktitle = {Proceedings of the IEEE Conference on Computer Vision and Pattern Recognition},
  pages     = {7132--7141},
  year      = {2018}
}

@inproceedings{woo2018cbam,
title={CBAM: Convolutional Block Attention Module},
author={Woo, Sanghyun and Park, Jongchan and Lee, Joon-Young and Kweon, In So},
booktitle={Proceedings of the European Conference on Computer Vision (ECCV)},
pages={3--19},
year={2018}
}

@inproceedings{wang2020eca,
title={ECA-Net: Efficient Channel Attention for Deep Convolutional Neural Networks},
author={Wang, Qilong and Wu, Banggu and Zhu, Pengfei and Li, Peihua and Zuo, Wangmeng and Hu, Qinghua},
booktitle={Proceedings of the IEEE/CVF Conference on Computer Vision and Pattern Recognition (CVPR)},
pages={11534--11542},
year={2020}
}

@inproceedings{li2019selective,
title={Selective Kernel Networks},
author={Li, Xiang and Wang, Wenhai and Hu, Xiaolin and Yang, Jian},
booktitle={Proceedings of the IEEE/CVF Conference on Computer Vision and Pattern Recognition (CVPR)},
pages={510--519},
year={2019}
}

@inproceedings{hou2021coordinate,
title={Coordinate Attention for Efficient Mobile Network Design},
author={Hou, Qibin and Zhou, Daquan and Feng, Jiashi},
booktitle={Proceedings of the IEEE/CVF Conference on Computer Vision and Pattern Recognition (CVPR)},
pages={13713--13722},
year={2021}
}

@misc{zhao2024rtdetr,
      title={DETRs Beat YOLOs on Real-time Object Detection},
      author={Wenyu Lv and Shangliang Xu and Yian Zhao and Guanzhong Wang and Jinman Wei and Cheng Cui and Yuning Du and Qingqing Dang and Yi Liu},
      year={2023},
      eprint={2304.08069},
      archivePrefix={arXiv},
      primaryClass={cs.CV}
}

@software{torchvision2016,
    title        = {TorchVision: PyTorch's Computer Vision library},
    author       = {TorchVision maintainers and contributors},
    year         = 2016,
    journal      = {GitHub repository},
    publisher    = {GitHub},
    howpublished = {\url{https://github.com/pytorch/vision}}
}

@article{Evangelidis2008ECC,
  title={Parametric Image Alignment Using Enhanced Correlation Coefficient Maximization},
  author={Evangelidis, Georgios D. and Psarakis, Emmanouil Z.},
  journal={IEEE Transactions on Pattern Analysis and Machine Intelligence},
  volume={30},
  number={10},
  pages={1858--1865},
  year={2008},
  doi={10.1109/TPAMI.2008.113}
}
\end{document}